\documentclass[10pt,twocolumn,letterpaper]{article}

\usepackage{cvpr}              

\usepackage{graphicx}
\usepackage{amsmath}
\usepackage{amssymb}
\usepackage{booktabs}
\usepackage{color}
\usepackage{multirow}

\newcommand\blfootnote[1]{%
	\begingroup 
	\renewcommand\thefootnote{}\footnote{#1}%
	\addtocounter{footnote}{-1}%
	\endgroup 
}

\usepackage[pagebackref,breaklinks,colorlinks]{hyperref}

\usepackage[capitalize]{cleveref}
\crefname{section}{Sec.}{Secs.}
\Crefname{section}{Section}{Sections}
\Crefname{table}{Table}{Tables}
\crefname{table}{Tab.}{Tabs.}

\def\cvprPaperID{143} 
\def\confName{CVPR}
\def\confYear{2023}

\begin{document}
	
	\title{NEF: Neural Edge Fields for 3D Parametric Curve Reconstruction\\from Multi-view Images}
	
	\author{
		Yunfan Ye \quad Renjiao Yi \quad Zhirui Gao \quad Chenyang Zhu \quad Zhiping Cai \quad Kai Xu\thanks{Corresponding author.} \\
		National University of Defense Technology
	}

	\twocolumn[{%
		\maketitle
		\centering
		\includegraphics[width=0.95\linewidth]{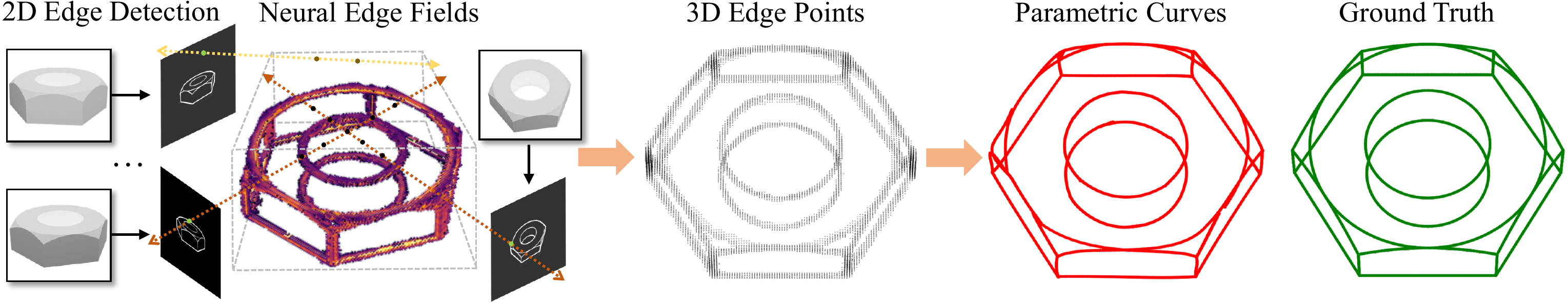}
		\vspace{-0.5em}
		\captionof{figure}{
			We leverage 2D edge detection to directly acquire 3D edge points by learning a neural implicit field and further reconstructing 3D parametric curves that represent the geometrical shape of the object. We introduce the details about extracting 3D edges from the proposed Neural Edge Field in~\ref{sec:extracting_3D_edges}, and the coarse-to-fine optimization strategy to reconstruct parametric curves in~\ref{sec:parametric_curve}. The whole pipeline is self-supervised with only 2D supervision.
		}
		\vspace{0.5em}
		\label{fig:teaser}
	}]
	
	\blfootnote{*Corresponding author.}
	
	\begin{abstract}
		\vspace{-0.3cm}
		We study the problem of reconstructing 3D feature curves of an object from a set of calibrated multi-view images. To do so, we learn a neural implicit field representing the density distribution of 3D edges which we refer to as Neural Edge Field (NEF). Inspired by NeRF~\cite{mildenhall2020nerf}, NEF is optimized with a view-based rendering loss where a 2D edge map is rendered at a given view and is compared to the ground-truth edge map extracted from the image of that view. The rendering-based differentiable optimization of NEF fully exploits 2D edge detection, without needing a supervision of 3D edges, a 3D geometric operator or cross-view edge correspondence. Several technical designs are devised to ensure learning a range-limited and view-independent NEF for robust edge extraction. The final parametric 3D curves are extracted from NEF with an iterative optimization method. On our benchmark with synthetic data, we demonstrate that NEF outperforms existing state-of-the-art methods on all metrics. Project page: \href{https://yunfan1202.github.io/NEF/}{https://yunfan1202.github.io/NEF/}.
		
	\end{abstract}
	

	\vspace{-0.2cm}
	\vspace{-0.4cm}
\section{Introduction}
\label{sec:intro}

Feature curves ``define'' 3D shapes to an extent, not only geometrically (surface reconstruction from curve networks~\cite{lengagne1996using,liu2008surface}) but also perceptually (feature curve based shape perception~\cite{decarlo2003suggestive,xu2009feature}). Therefore, feature curve extraction has been a long-standing problem in both graphics and vision. Traditional approaches to 3D curve extraction often work directly on 3D shapes represented by, e.g., polygonal meshes or point clouds.
Such approaches come with a major difficulty: Sharp edges may be partly broken or completely missed due to imperfect 3D acquisition and/or reconstruction. Consequently, geometrically-based methods, even the state-of-the-art ones, are sensitive to parameter settings and error-prune near rounded edges, noise, and sparse data. Recently, learning-based methods are proposed to address these issues but with limited generality~\cite{yu2018ec,wang2020pie,matveev2022def, liu2021pc2wf}.

In many cases, edges are visually prominent and easy to detect in the 2D images of a 3D shape. To resolve occlusion, one may think of 3D curve reconstruction from multi-view edges. This solution, however, relies strongly on cross-view edge correspondence which itself is a highly difficult problem~\cite{sinha2005multi}. This explains why there is rarely a work on multi-view curve reconstruction even in the deep learning era. We ask this question: Can we learn 3D feature curve extraction directly from the input of multi-view images?

In this work, we try to answer the question through learning a neural implicit field representing the density distribution of 3D edges from a set of calibrated multi-view images, inspired by the recent success of neural radiance field (NeRF)~\cite{mildenhall2020nerf}. We refer to this edge density field as Neural Edge Field (NEF). Similar to NeRF, NEF is optimized with a view-based rendering loss where a 2D edge map is rendered at a given view and is compared to the ground-truth edge map extracted from the image of that view. The volumetric rendering is based on edge density and color (gray-scale) predicted by MLPs along viewing rays. Different from NeRF, however, our goal is merely to optimize the NEF which is later used for extracting parametric 3D curves; no novel view synthesis is involved. The rendering-based differentiable optimization of 3D edge density fully exploits 2D edge detection, without needing a 3D geometric operator or cross-view edge correspondence. The latter is implicitly learned with multi-view consistency.

Directly optimizing NEF as NeRF-like density is problematic since the range of density can be arbitrarily large and different from scene to scene, and it is hard to select a proper threshold to extract useful geometric shapes (e.g., 3D surfaces for NeRF and 3D edges for NEF). Moreover, NeRF density usually does not approximate the underlying 3D shape well due to noise. Therefore, we seek for confining the edge density in the range of $[0,1]$ through learning a mapping function with a learnable scaling factor to map the edge density to the actual NEF density. By doing so, we can easily choose a threshold to extract edges robustly from the optimized edge density.

Another issue with NEF optimization is the incompatible visibility of the edge density field and the edges detected in images. While the former is basically a wireframe representation of the underlying 3D shape and all edges are visible from any view (i.e., no self-occlusion), edges in 2D images can be occluded by the object itself. This leads to inconsistent supervisions of different views with different visibility and may cause false negative: An edge that should have been present in NEF according to one view visible to the edge may be suppressed by other views invisible. To address this issue, we opt to 1) impose consistency between density and color in NEF and 2) give less punishment on non-edge pixels in the rendering loss, to allow the NEF to keep all edges seen from all views. This essentially makes NEF view-independent which is reasonable.

Having obtained the edge density, we fit parametric curves by treating the 3D density volume as a point cloud of edges. We optimize the control points of curves in a coarse-to-fine manner. Since initialization is highly important to such a non-convex optimization, we first apply line fitting in a greedy fashion to cover most points. Based on the initialization, we then upgrade lines to cubic Bézier curves by adding extra control points and optimize all curves simultaneously with an extra endpoint regularization.

We build a benchmark with a synthetic dataset consisting of 115 CAD models with complicated shape structures from ABC dataset~\cite{koch2019abc} and utilize BlenderProc~\cite{blenderproc} to render posed images. Extensive experiments on the proposed dataset show that NEF, which is self-trained with only 2D supervisions, outperforms existing state-of-the-art 
methods on all metrics. Our contributions include:
\begin{itemize}
	\vspace{-5pt}\item A self-supervised 3D edge detection from multi-view 2D edges based neural implicit field optimization.
	
	\vspace{-5pt}\item Several technical designs to ensure learning a range-limited and view-independent NEF and an iterative optimization strategy to reconstruct parametric curves.
	\vspace{-5pt}\item A benchmark for evaluating and comparing various edge/curve extraction methods.
\end{itemize}
	\section{Related Work}
\label{sec:related}

\paragraph{Neural Radiance Fields.} NeRF~\cite{mildenhall2020nerf} have demonstrated
the remarkable ability for novel view synthesis. 
The basic idea of NeRF to represent the geometry and appearance of a scene as a radiance field, allowing querying color and volume density in continuous spatial positions and viewing directions for rendering. Many extensions are designed on the NeRF backbone, such as speeding up the training~\cite{reiser2021kilonerf, sun2022direct} and the inference~\cite{hedman2021baking, yu2021plenoctrees, chen2022mobilenerf}, editing~\cite{liu2021editing, yuan2022nerf, wang2022clip}, generative models~\cite{schwarz2020graf, niemeyer2021giraffe} and model reconstruction~\cite{yariv2021volume, wang2021neus, oechsle2021unisurf}, and more are discussed in ~\cite{dellaert2020neural, gao2022nerf}. However, there are not many works that utilize NeRFs to extract 3D skeletons/curves. We propose Neural Edge Fields (NEF) to reconstruct 3D edges from 2D images. 
The closest NeRF-based works to ours are for model reconstruction~\cite{yariv2021volume, wang2021neus}, and we all recover the precise shape geometry by defining the original density as a transformed new representation. The difference is, they represent the surface by zero-level sets of the signed distance function (SDF) and focus on surface reconstruction; while we introduce edge density to represent the edge probability at every spatial position by learning a NEF.
 
\vspace{-0.2cm}
\paragraph{3D Parametric Curve Reconstruction.} The basis of 3D parametric curve reconstruction is point cloud edge detection algorithms. 
Traditional (non-learning) methods focus on multi-view images~\cite{prakoonwit20073d}, or local geometric properties of point clouds such as normals~\cite{demarsin2007detection, weber2010sharp}, curvatures~\cite{yang2014automated}, and hierarchical clustering~\cite{feng2014fast}. Recent data-driven methods often adopt edge detection as a binary classification for point clouds. For each possible edge point, its neighborhood attributes are taken as the learning features. With the progress of network architectures, the classifier for edge detection ranges from random forests~\cite{hackel2016contour, hackel2017joint}, Pointwise Multilayer Perceptron (MLP)~\cite{yu2018ec, wang2020pie} based on PointNet++~\cite{qi2017pointnet++}, to capsule networks~\cite{bazazian2021edc}.

Representing point cloud edges as parametric curves is more challenging.
PIE-NET~\cite{wang2020pie} learns to detect edges and corners from point clouds, and generates parametric curve proposals using networks, suppressing the invalid ones at last. 
PC2WF~\cite{liu2021pc2wf} is composed of a sequence of feed-forward blocks to sample point cloud as patches, to classify if the patch contains a corner. They regress the location of corners, and connect all corners as parametric curves. 
DEF~\cite{matveev2022def} calculates estimates of the truncated distance-to-feature field for each input point cloud by an extra set of depth images in a patch-based manner, and fitting curves after corner detection and clustering. 
Unlike those works requiring at least point clouds as input and training from labeled datasets, our method is self-supervised by 2D 
edges.

\section{Method}

\label{sec:method}
To obtain 3D parametric curves from multi-view images, the method consists of two steps: building neural edge fields (NEF) and reconstructing parametric curves. As illustrated in Fig.~\ref{fig:teaser}, 2D edge maps are predicted by a state-of-the-art edge detection network PiDiNet~\cite{su2021pixel}, and NEF is built from these multi-view edge maps, as introduced in Sec.~\ref{sec:extracting_3D_edges}. 
Adopting NeRF directly on edge maps is problematic, and there are many differences between edge maps and original images. Edges are sparse, and inconsistent among views due to occlusions. To deal with them, We introduce several training losses specifically designed for NEF.  
To reconstruct the parametric curves from 3D edge points, we introduce two-stage coarse-to-fine optimization in Sec.~\ref{sec:parametric_curve}. 
In the coarse-stage, we simplify curves to straight lines, and fit a group of lines to 3D edge points in a fit-and-delete strategy. In the fine-stage, we upgrade straight lines into cubic Bézier curves by adding extra control points.  

\subsection{Reconstructing 3D Edge Points}
\label{sec:extracting_3D_edges}
In this section, we learn a neural implicit filed representing the spatial distribution of 3D edges, named neural edge field (NEF). 
We first introduce preliminaries about NeRF in Sec.~\ref{sec:Preliminaries}. The design of NEF is introduced in Sec.~\ref{sec:NEF}. Training NEF requires specific loss designs, as introduced in Sec.~\ref{sec:loss_functions}. 

\vspace{-0.3cm}
\subsubsection{Preliminaries}
\label{sec:Preliminaries}
NeRF~\cite{mildenhall2020nerf} represents a continuous scene with an MLP network, which maps 5D coordinates (location $(x, y, z)$ and viewing direction $(\theta, \phi)$) among camera rays, to color $(r, g, b)$ and volume density $\sigma$. After training, novel views can be rendered from arbitrary camera poses, following volume rendering. 
Given the camera origin $o$ and ray direction $d$ with near and far bounds $t_{n}$ and $t_{f}$, the predicted pixel color $\hat{C}$ of camera ray $r(t) = o+td$ is defined as follows:
\begin{equation}
	\hat{C}({r})=\int_{t_{n}}^{t_{f}} T(t) \sigma({r}(t)) {c}({r}(t), {d}) d t,
\end{equation}
where $T(t)=\exp \left(-\int_{t_{n}}^{t} \sigma({r}(s)) d s\right)$, and densities $\sigma$ and colors $c$ are predictions of the MLP network. The loss function of NeRF is a re-rendering loss defined by the mean squared error (MSE) between the rendered color $\hat{C}$ and the ground truth pixel color $C$: 
\begin{equation}
	\mathcal{L}_{color} = \sum_{r \in R_{i}}|| (C(r) - \hat{C}(r)) ||^2,
\end{equation}
where $R_i$ is the set of rays in each training batch.

In our scenario, we adopt the structure of NeRF as backbone, but modify the color $c=(r, g, b)$ to one-dimensional gray value as $c=(gray)$ to represent edge intensities.

\vspace{-0.2cm}
\subsubsection{Neural Edge Fields} \label{sec:NEF}
\begin{figure}
	\centering
	\begin{subfigure}{0.49\linewidth}
		\includegraphics[width=0.998\columnwidth]{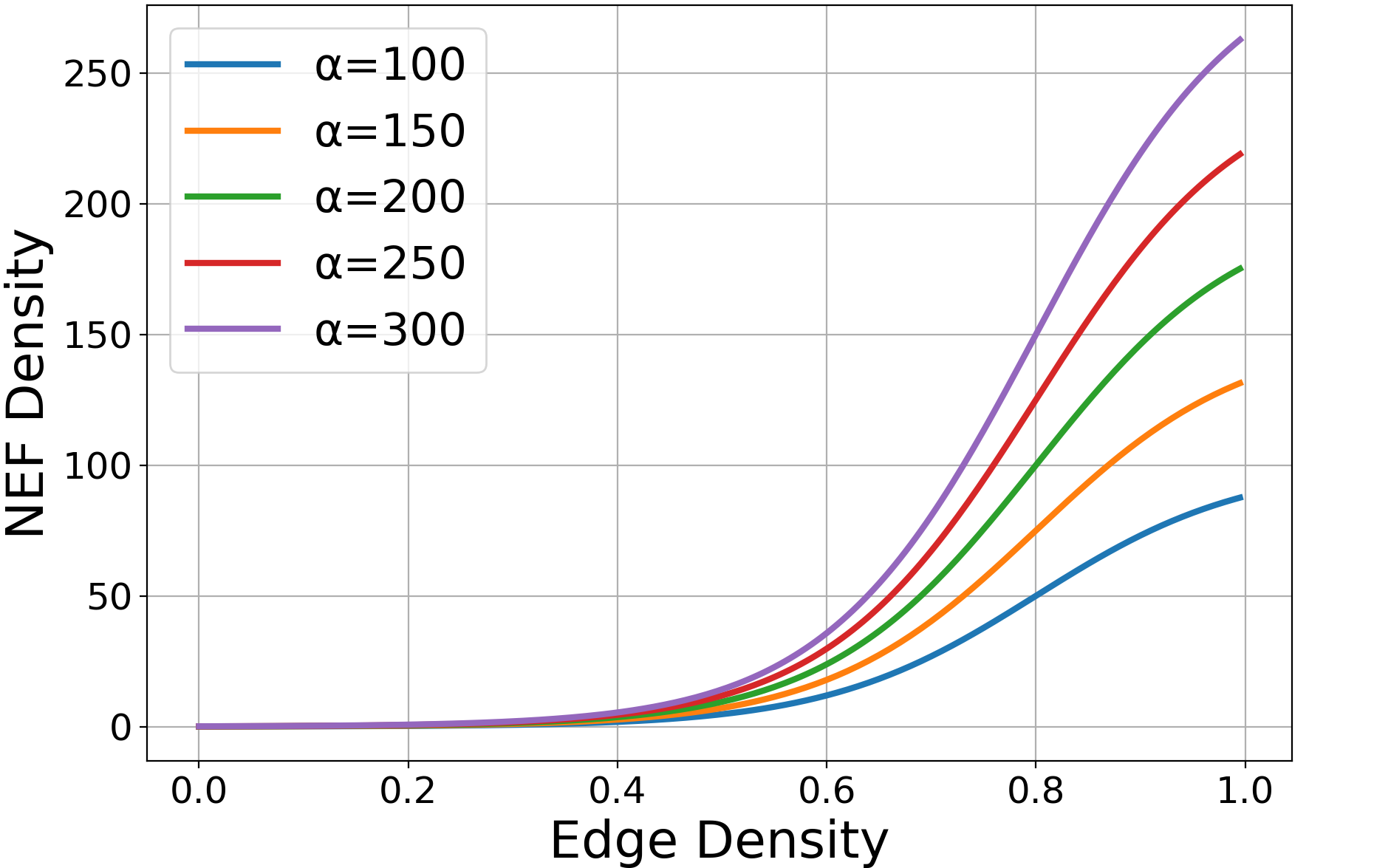}
		\caption{The curves of the transformation from edge density to the volume density with different $\alpha$.}
		\label{fig:alpha-a}
	\end{subfigure}
	\hfill
	\begin{subfigure}{0.49\linewidth}
		\includegraphics[width=1\columnwidth]{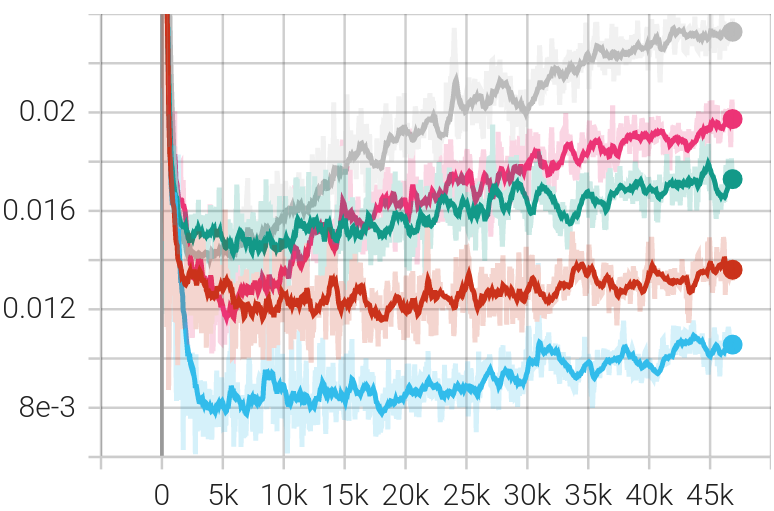}
		\caption{The curves of five randomly selected samples that show $\alpha\times10^{-4}$ during training iterations.}
		\label{fig:alpha-b}
	\end{subfigure}
	\caption{Examples of the transformation and variety trend of $\alpha$ in Eqn~\ref{eq:edge_density}. As in (a), with edge density ranging from 0 to 1, we can adaptively match NEF density. 
		Adapting to different scenes is controlled by the trainable parameter $\alpha$ as in (b).} 
	
	\label{fig:alpha_curves}
\end{figure}

\begin{figure}[t]
	\centering
	\includegraphics[width=0.95\linewidth]{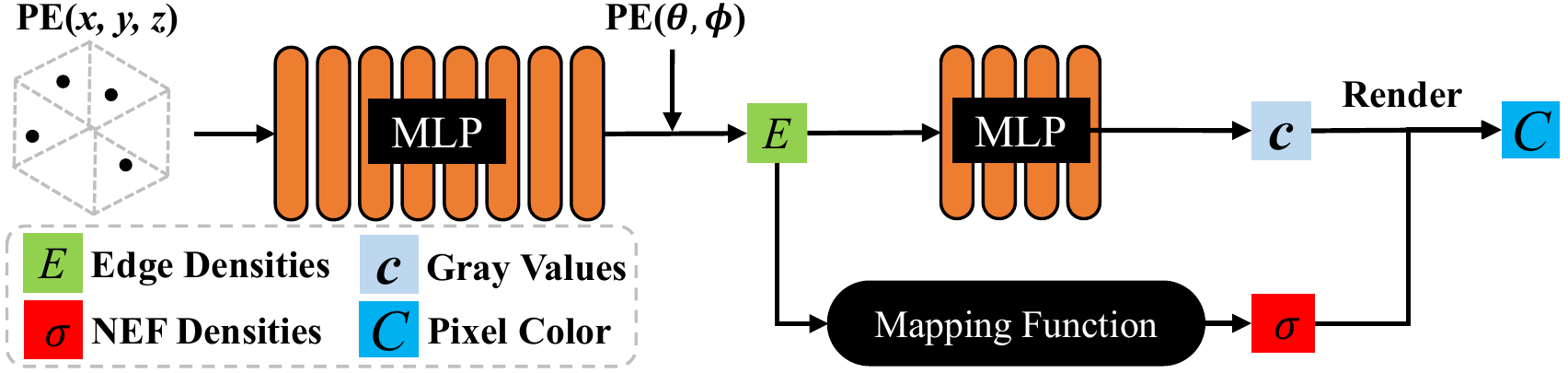}
	\caption{3D location $(x, y, z)$ and viewing direction $(\theta, \phi)$ are fed into the network after positional encoding (PE). The NEF density $\sigma$ is mapped from the edge density with a learnable scale.}
	\label{fig:network}
\end{figure}

We introduce neural edge fields (NEF), training from 2D edge maps to represent the edge probability at every spatial positions. 
While NeRFs synthesize photorealistic novel views images based on the differentiable volume rendering, their volume density does not approximate the actual 3D shape very well. Similar cases also exist for NEF. The NEF density does not approximate the actual 3D edges. 
The range and scale of NEF density also varies from scene to scene, making the 3D edges difficult to extract from them. 
Recently,  NeuS~\cite{wang2021neus} and VolSDF~\cite{yariv2021volume} represent the object surface by the signed distance function (SDF) and mapping SDF to volume density of NeRFs by a distribution function. 
Similarly, we introduce an intermediate density field, called edge density, before NEF densities, as illustrated in Fig.~\ref{fig:network}. 
During training, with proper supervisions/constraints on edge densities, they are expected to approximate the 3D edges well. Edge density describes the edge probability at each position. It is in the range of $[0, 1]$, to be unified with 2D edges ($1$ represents edge and $0$ is non-edge).
After mapping functions, we can transform edge densities to NEF densities, which are used for volume rendering. Given ${\bf x} \in \mathbb{R}^3$ represent the space occupied by the object in ${R}^3$, and $E(x)$ represent the value of edge density in location $\bf x$, the NEF density $\sigma$ is calculated by:
\begin{equation}\label{eq:edge_density}
	\sigma(x) = \alpha (1 + e^{-g(E(x)-\beta)})^{-1},
\end{equation} 
where $\alpha$ is a trainable parameter to control the density scale, $\beta$ is the mean to control the function position, and $g$ adjust the distribution around $[0, 1]$. The edge density is expected to adaptively match the distribution of NEF density, and should also be easily binarized by a unified threshold. Thus, to ensure a proper mapping from edge density to NEF density, we set $\beta=0.8$ and $g=10$ in all experiments, and $\alpha$ is a trainable parameter. As illustrated in Fig~\ref{fig:alpha_curves}, $\alpha$ value varies from scene to scene. The edge density is obtained by adding an extra sigmoid layer after the original NeRF MLPs. We add another MLPs of 4 hidden layers with a size of 256 to predict the gray value $c$ by edge densities and view directions. The network architecture is shown in Fig.~\ref{fig:network}.

\vspace{-0.2cm}
\subsubsection{Training NEFs} \label{sec:loss_functions}

Training NEF as NeRF is problematic due to several problems. Firstly, 3D edges are similar to 3D skeletons of the object, and in volume rendering, samples are sparse along rays, making the network easily stuck in local optima. Secondly, 2D edge maps do not match the actual 3D wireframe, due to occlusions. Edges may not be visible in all views, leading to inconsistencies among views. We introduce the weighted mean squared error loss (W-MSE) and consistency loss to solve these two problems. Furthermore, to encourage the sparsity of points in NEF, we also introduce a sparsity loss. 

With three loss designs, we are able to stably train NEF supervised by 2D images. 
The final loss function is represented as:
\begin{equation}
	\mathcal{L} = \lambda_{1}\mathcal{L}_{color} + \lambda_{2}\mathcal{L}_{consistency} + \lambda_{3}\mathcal{L}_{sparsity},
\end{equation}
where the balancing parameters $\lambda_{1}$, $\lambda_{2}$, $\lambda_{3}$ are set to 1, 1 and 0.01 respectively in all experiments. 

Once NEF is trained, we set a fixed threshold of $0.7$ to binarize edge densities, to extract 3D edge points from NEF. 

\vspace{-0.2cm}
\paragraph{W-MSE loss.}
We obtain 2D edge maps from a lightweight edge detector PiDiNet~\cite{su2021pixel}. On edge maps, edge pixels are in white color and often sparsely distributed. Therefore, when training NEF, edge and non-edge pixels are highly imbalanced, which leads to very sparse samples along rays. In this case, the network is easy to degenerate to local optima. A most common degeneration case is to predict all densities and colors are zeros, and the rendered images are all black. Therefore, we modify the original color loss by adding an adaptive weight $W(r)$ in each batch. The weighted mean squared error loss (W-MSE) is defined as:
\begin{equation}\label{eq:wmse_loss}
	\mathcal{L}_{color} = \sum_{r \in R_{i}} W(r)|| (C(r) - \hat{C}(r)) ||^2,
\end{equation}
in which 
\begin{equation}
	W(r)=
	\left\{
	\begin{array}{lll}
		\frac{\left|C^{+}\right|}{\left|C^{+}\right|+\left|C^{-}\right|}, &  if \ C(r)<=\eta,\\
		\frac{\left|C^{-}\right|}{|C^{+}|+|C^{-}|}, & if \ C(r)>\eta,
	\end{array}
	\right.
\end{equation} 
where $C^+$ and $C^-$ denote the number of edge and non-edge pixels in each batch decided by the threshold $\eta$. We set $\eta$ to 0.3 throughout the paper. The adaptive weight is simple yet effective by enforcing the network to focus more on edge pixels/rays, avoiding degenerating. 

\vspace{-0.3cm}
\paragraph{Consistency loss.}
The edge map of each view does not match the real 3D wireframe. On a 2D edge map, not all edges are visible due to occlusions. It means the ``ground truths" are not exactly correct, missing those invisible edges in each view. 

\begin{figure}[t]
	\centering
	\includegraphics[width=0.95\linewidth]{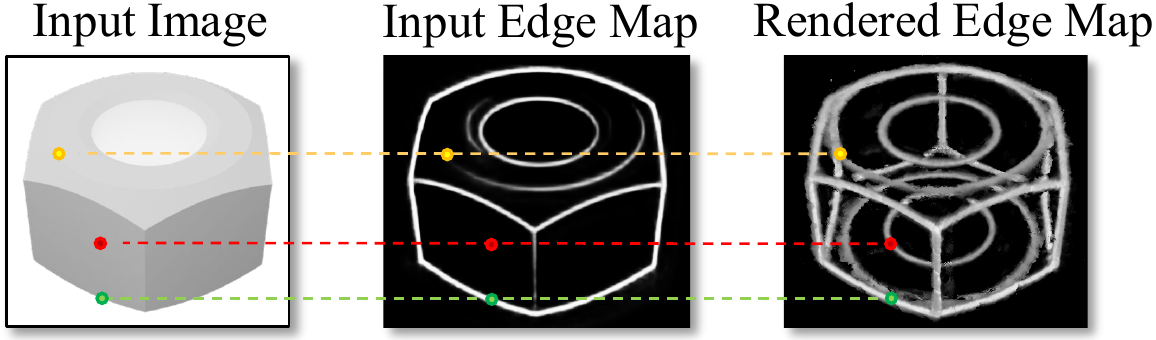}
	\caption{The green points denotes the edges that can be seen and detected in a given view. The yellow points mean edges that are visible in this view, but are hard to be detected (i.e. improper illumination). The red ones mean edges that are totally occluded but can be seen in other views. Our method integrates edges seen from multiple views, and can re-render all edges for this view.}
	\label{fig:occluded_edge}
\end{figure}

To successfully reconstruct 3D edge points from these 2D edge maps, 
we should recover the complete edge map of each view by integrating the information from other views where the occluded edges are visible, as shown in Fig~\ref{fig:occluded_edge}. 
For each view, occluded edges are invisible in the image, as well as the edge map. 
NEF will be confused during training due to such inconsistencies among views. 
In each view, there are many false-negative pixels which on ``ground truth'' are non-edges. 
Such inconsistency brings noisy NEF around the object surface in spatial positions. For these occluded edges, the value of edge density is close to $1$, but the color is close to $0$ to fit those false-negative samples. 
Therefore, we enforce the value of edge density and color intensity (both are within $[0, 1]$) to be consistent for all samples along rays, to reduce the false-negative pixels. The consistency loss is also calculated by mean squared error, and defined as:
\begin{equation}\label{eq:consistency}
	\mathcal{L}_{consistency} = \sum_{r \in R_{i}}|| (E(r) - {c}(r)) ||^2.
\end{equation}

Since the W-MSE loss in Eqn.~\ref{eq:wmse_loss} encourages NEF to focus more on edge pixels and give less punishment to non-edge pixels. Thus, combining W-MSE loss and consistency loss, not only stabilizes the training process, but also encourages NEF to occluded edges, by learning from other views. After training, the re-rendered edge maps by NEF successfully recover those invisible edges, as illustrated in Fig.~\ref{fig:occluded_edge}. Thus, adopting different 2D edge detectors have limited influences on NEF reconstruction. With the consistency loss, NEF automatically corrects missing 2D edges, no matter they are occluded or missed by detectors (See Sec.~\ref{sec:ablation}). 

\vspace{-0.3cm}
\paragraph{Sparsity loss.}
As mentioned, edges are sparse in both 2D and 3D spaces. To encourage spatial sparsity, 
as well as accelerating the convergence, we add an extra regularizer, sparsity loss, to penalize unnecessary edge densities along rays of non-edge pixels during training. 
We adopt Cauchy loss~\cite{barron2019general} as the sparsity regularizer, which is highly robust to outliers. The sparsity loss is defined as:
\begin{equation}
	\mathcal{L}_{sparsity} = \sum_{i, j} \text{log}\left(1 + \frac{E(r_i (t_j))^2}{s} \right) \, ,
\end{equation}
where $i$ indexes non-edge pixels of input edge maps, $j$ indexes samples along the corresponding rays, $s$ control the scale of the regularizer. We fix $s=0.5$.

\subsection{Extracting 3D Parametric Curves}
\label{sec:parametric_curve}
With 3D edge point clouds from NEF, we further extract parametric curves. 
We extract Bézier curves from 3D edges in a coarse-to-fine manner. The objective of optimization is introduced in Sec.~\ref{sec:optimization_loss}. The coarse-to-fine pipeline is introduced in Sec.~\ref{sec:coarsetofine}.


\vspace{-0.3cm}
\subsubsection{Bézier Curves Optimization.}
\label{sec:optimization_loss}
We adopt cubic Bézier curves to represent the geometrical shape of 3D edges. For each curve, we optimize the positions of four control points. The first and last control points define the beginning and ending positions, and the other two control points lead to different curvatures. Straight lines can be considered as linear Bézier curves with two control points. 
The goal is to optimize a set of parameters (positions of four control points) for all curves $\{curve_i\}_{i=1}^{n} = \{\{p_i^j\}_{j=1}^4\}_{i=1}^n$
to fit the 3D point cloud. The number of curves $n$ varies for different objects. 
To optimize the curve fitting, we sample 100 points on each curve, and dilate them up to 500 by adding Gaussian noise around them. 
We apply the widely-used Chamfer Distance (CD) to compute the distances between the curve points and 3D edge points. $P_c$ and $P_t$ represent the sampled points from curves and the target 3D edge points, respectively, the Chamfer loss is defined as: 
\begin{equation}
	\begin{aligned}
		\mathcal{L}_{CD}(P_c, P_t) = \gamma &\frac{1}{|P_c|}\sum_{x \in P_c} \min_{y \in P_t} \| x-y \|_{2}^{2} + \\ &\frac{1}{|P_t|}\sum_{y \in P_t} \min_{x \in P_c} \| x-y \|_{2}^{2},
		\label{eq:chamfer_distance}
	\end{aligned}	
\end{equation}
where $\gamma$ is the parameter to control the tendency for each side ($\gamma=1$ for original Chamfer loss). Each point in $P_c$ finds the closest point in $P_t$ (and vice-versa), and calculates the average pair-wise point-level distance. A bigger $\gamma$ means the optimization focus more on minimizing the distance from $P_c$ to $P_t$. 

\begin{figure}
	\centering
	\includegraphics[width=0.95\linewidth]{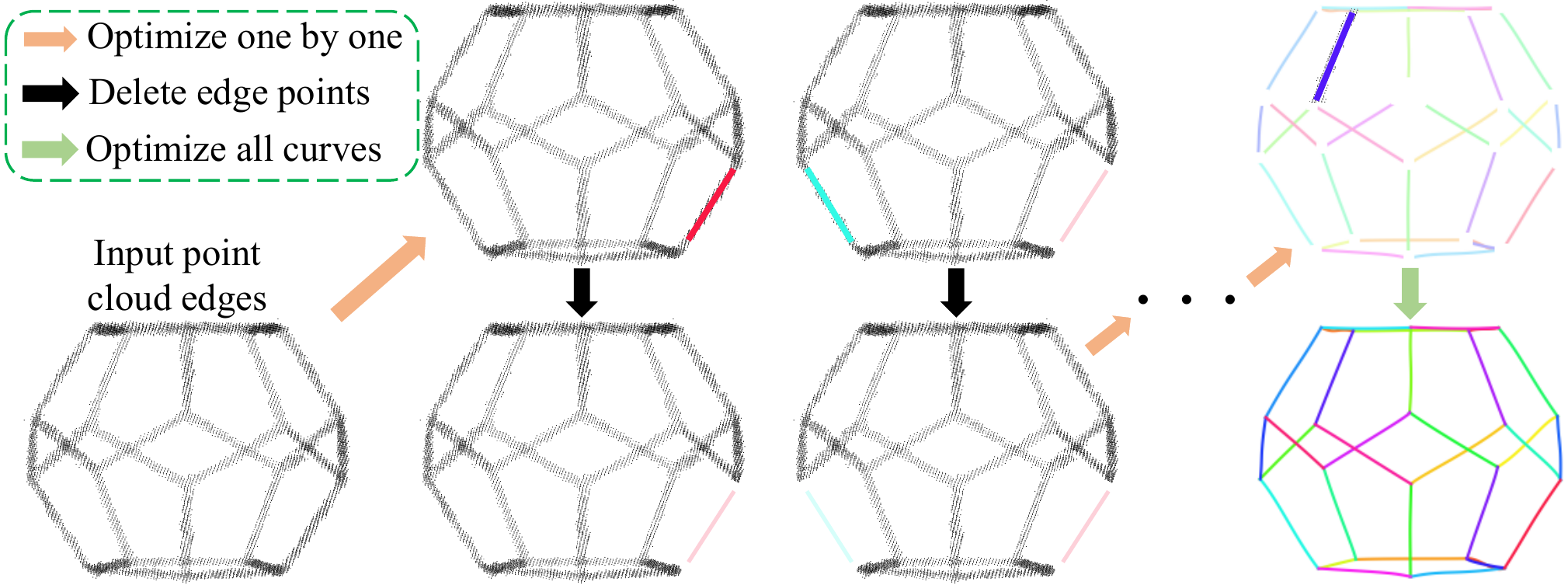}
	\caption{We iteratively optimize lines one by one to fit the 3D edge points, and follow a fit-and-delete strategy. The process continues until very few points are left. Fitted lines are shown with different colors.}
	\label{fig:iterative}
\end{figure}

By minimizing CD, we fit Bézier curves to 3D edge points. However, the optimization of CD is insensitive to endpoint details, and we find that many curves are not connected. 
To encourage curve connections, we add a regularizer in the objective function, to encourage endpoints which are close in space to meet. 
Two endpoints of each Bézier curve are the first and the last control points, and the endpoints of all curves $\{curve_i\}_{i=1}^{n}$ are termed as $P_E = \{\{p_i^j\}_{j=1, 4}\}_{i=1}^n$. The endpoint regularizer is defined as:
\begin{equation}\label{eq:endpoints_loss}
	\mathcal{L}_{EP} = \sum_{x,y \in P_E} M\| x-y \|_{2}^{2},
\end{equation}
in which
\begin{equation}\label{eq:endpoint_mask}
	M =
	\left\{
	\begin{array}{lll}
		1, &  if \ x-y<=d,\\
		0, & if \ x-y>d,
	\end{array}
	\right.
\end{equation} 
$M$ is a mask to ensure the endpoints loss only regularize those endpoints that are close enough to each other (within the distance $d$). 
At last, the objective function to optimize all curves:
\begin{equation}
	\underset{\{\{p_i^j\}_{j=1}^4\}_{i=1}^n}{\arg\min}(\mathcal{L}_{CD} + \lambda\mathcal{L}_{EP}),
	\label{eq:curve_optimization}
\end{equation}
where $\lambda$ is set to 0.01 in all experiments.

\begin{figure*}[t]
	\centering
	\includegraphics[width=0.87\linewidth]{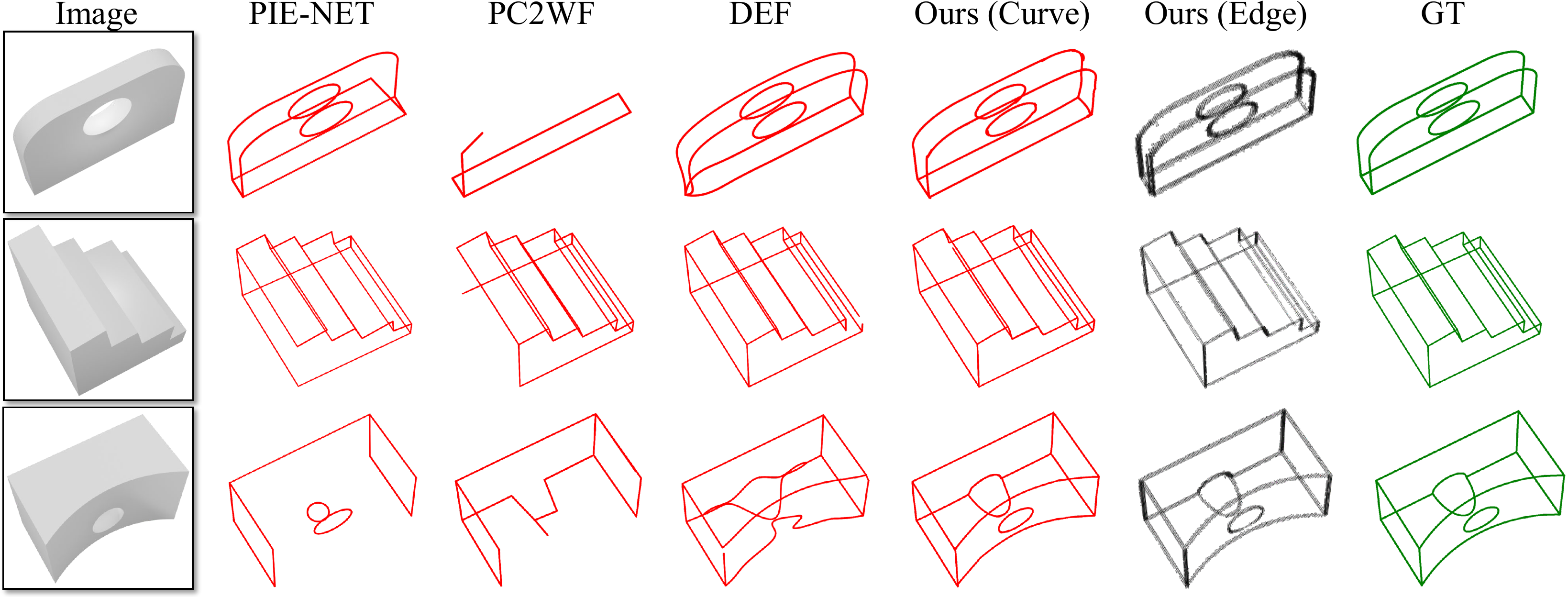}
	\caption{Qualitative comparisons against other methods. From left to right, we present the rendered image, the results of PIE-NET, PC2WF, DEF, our reconstructed curves, our point cloud edges obtained from edge densities, and the ground truth edges. Other approaches are trained on point clouds from ABC dataset, and ours is self-supervised by 2D edge maps.}
	\label{fig:comparisons}
\end{figure*}

\vspace{-0.2cm}
\subsubsection{Coarse-to-fine Scheme}
\label{sec:coarsetofine}

The objective function of optimization is
highly non-convex, making it easy for control points to converge to local minima. Therefore, the initialization of Bézier curves has a significant impact on the final result of the optimization. It is also difficult to select a proper number of curves that is suitable for all objects. Thus, we design a coarse-to-fine pipeline to extract curves. At coarse-level, we downgrade cubic Bézier to straight lines, and fit a set of lines to 3D edge points. At fine-level, we upgrade lines to cubic Bézier curves, and connect the endpoints of curves. 

\vspace{-0.2cm}
\paragraph{Coarse-level optimization. }

Instead of optimizing all lines simultaneously, we iteratively optimize the lines one by one with a fit-and-delete strategy.
Specifically, during each iteration, we enlarge $\gamma$ to 5 in Chamfer loss (Eqn.~\ref{eq:chamfer_distance}), to encourage one line (linear curve) to fit partial 3D edge points to the utmost. After one line is decided, we delete the 3D edge points around the line and record its parameters. The process continues until there are not many 3D edge points left (i.e. $<$20). 
The group of lines fitted works as initialization to the fine level. 
We demonstrate the process of coarse-level optimization in Fig~\ref{fig:iterative}. Since the lines are fitted one by one, we do not consider endpoints regularization at this level. 

\vspace{-0.2cm}
\paragraph{Fine-level optimization. }Coarse-level optimization initializes the number of lines, as well as the beginning and ending positions. In fine-level, we upgrade all straight lines back to cubic Bézier curves by interpolating another two control points between the endpoints pair, and solve the optimization in Eqn.~\ref{eq:curve_optimization}. The resulting parametric curves precisely match the 3D edge points as demonstrated in Fig.~\ref{fig:comparisons}. 


\begin{figure*}[t]
	\centering
	\includegraphics[width=0.9\linewidth]{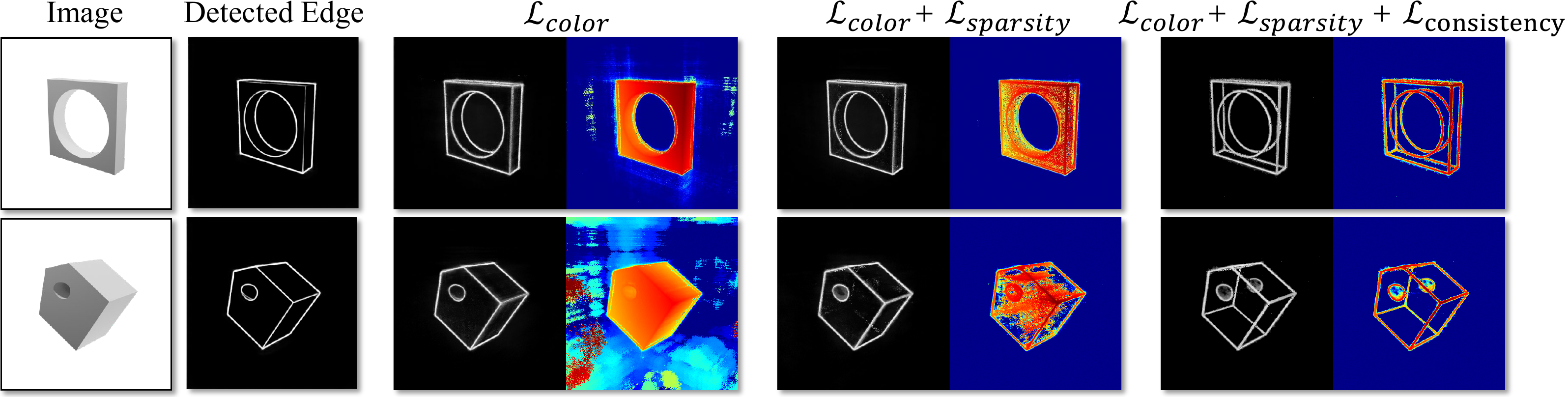}
	\caption{From left to right, we present 2D images and detected edge maps in a given view, followed by the rendered edge map and depth map of three loss combinations. Rendered depth maps show the spatial distribution of the edge density field. The sparsity regularization eliminates most noise around the object, and the consistency loss makes the edge density clearly aligned with 2D edges and easy to be separated from the background.}
	\label{fig:ablation_nerf}
	\vspace{-0.2cm}
\end{figure*}

\section{Experiments}

We compare with state-of-the-art methods and conduct ablations on the contributed ABC-NEF dataset. More experiments, discussions, training details and video demos are in the Supplementary. 
\subsection{ABC-NEF Dataset.}
As previous works~\cite{wang2020pie, liu2021pc2wf, matveev2022def}, we conduct experiments on the ABC dataset~\cite{koch2019abc} which consists of more than one million CAD models with edge annotations. To evaluate our pipeline, we provide a dataset called ABC-NEF, consisting of 115 distinct and challenging CAD models. They include various types of surfaces and curves, from the first chunk of the ABC dataset. We adopt BlenderProc~\cite{blenderproc} to render posed images facing the center of the object. We sample 50 views of $800\times800$ image rays for each object. The 50 views are sampled by evenly placing cameras on a sphere by Fibonacci sampling~\cite{hannay2004fibonacci}. 
Statistical analysis of ABC-NEF dataset and ablations about the number of views are included in the supplementary material.


\subsection{Comparisons with state-of-the-arts}
\paragraph{Comparison Settings.}
We compare the proposed method with three most state-of-the-art data-driven methods of parametric curve reconstruction, PIE-NET~\cite{wang2020pie}, PC2WF~\cite{liu2021pc2wf} and DEF~\cite{matveev2022def}. All three methods require point clouds as inputs, while ours require only 2D images. 
Following the settings in their papers, for PIE-NET, we apply the farthest point sampling method to uniformly sample 8096 clean point clouds that represent the object shape as the inputs, the outputs of PIE-NET contains closed and open curves. 
For PC2WF, we sub-sample 200,000 points for each object from surface meshes as the inputs, it outputs pairs of endpoints representing straight lines. For DEF, the inputs contain 128 views of depth maps and point clouds. In DEF, depth maps are collected to build a distance-to-feature field, which is used to detect corners on point clouds and extract spline curves.

We adopt their pre-trained models to reconstruct parametric curves for evaluation. 
Since PC2WF is designed to detect straight lines, we also make comparisons on a subset of the proposed ABC-NEF which contains 26 CAD models of only lines, named ABC-NEF-Line. 

\vspace{-0.3cm}
\paragraph{Evaluation Metrics.} We sample points on reconstructed parametric curves and evaluate distances between the sampled points and ground truth edge points. To ensure the points are evenly distributed, we down-sample the points on voxel grids, so there is at most one point per voxel.

To measure the location of reconstructed 3D edges, we adopt the Intersection over Union (IoU), precision, recall and their F-score. However, a small shift between two point clouds may lead to large changes on above metrics. We also adopt the Chamfer Distance (CD) between point clouds, to measure the geometric accuracy of reconstructed parametric curves. A small shift between point clouds would not affect much for CD. Before comparison, we normalize and align all ground truth edges and curve predictions into the range of $[0, 1]$. After normalization, when evaluating IoU, precision, recall and F-score, points are considered matched if there exists at least one ground truth point with the L2 distance smaller than 0.02.


\vspace{-0.3cm}
\paragraph{Comparison Results.} As reported in Table~\ref{tab:comparison}, our self-supervised method with only 2D supervisions significantly outperforms other state-of-the-arts in all metrics and datasets. 
We observe that PIE-NET and PC2WF both achieve much higher precision than recall, which means they often miss curves, but the detected curves locate precisely. 
Although PC2WF is designed to detect straight lines, we notice on ABC-NEF-Line dataset, our method still achieves better performance.

We illustrate qualitative performance in Fig.~\ref{fig:comparisons}. The results show that PIE-NET and DEF can detect and locate most curves well, and PC2WF is proficient in reconstructing line structures. However, limited by the design, PC2WF can only detect lines and struggles in capturing any other types of curves. Since PIE-NET is exactly trained on sharp features, leading to incompetent performance around ellipse edges and areas with relatively weak curvature. Meanwhile, DEF reconstructs curves mainly based on a continuous and smooth distance-to-feature field, thus has trouble in discriminating close curves and tends to incorrectly connect adjacent curves. 
We also notice these methods reconstruct curves heavily relying on corner detection, thus failing to cover all edges if some corners are missed. 
Essentially, these data-driven methods, may suffer from reconstructing curves for out-of-distribution shapes. On the contrary, our method benefits from the self-supervised pipeline, and can be trained on natural images. 
More comparisons and discussions are in the supplementary material.

\begin{table}
	\centering
	\scalebox{0.75}{
		\begin{tabular}{c|c|ccccc}
			\hline
			Dataset & Method & CD↓   & precision↑ & recall↑ & F-score↑ & IoU↑ \\
			\hline
			\multirow{4}[2]{*}{A-N} & PIE-NET & 0.0708 & 0.9072 & 0.7204 & 0.7846 & 0.6709 \\
			& PC2WF & 0.1382 & 0.9043 & 0.5525 & 0.6348 & 0.5074 \\
			& DEF   & 0.0402 & 0.8343 & 0.7802 & 0.8009 & 0.7368 \\
			& Ours  & \textbf{0.0353} & \textbf{0.9387} & \textbf{0.8838} & \textbf{0.9044} & \textbf{0.8283} \\
			\hline
			\multirow{4}[2]{*}{A-N-L} & PIE-NET & 0.0409 & 0.9481 & 0.8321 & 0.8803 & 0.7934 \\
			& PC2WF & 0.0614 & 0.9317 & 0.7746 & 0.8200  & 0.7492 \\
			& DEF& 0.0433 & 0.8118 & 0.7551 & 0.7757 & 0.7197 \\
			& Ours  & \textbf{0.0287} & \textbf{0.9717} & \textbf{0.9070} & \textbf{0.9353} & \textbf{0.8766} \\
			\hline
	\end{tabular}}%
	\caption{Quantitative comparisons to state-of-the-art methods. Note that our method is self-supervised by 2D 
	edge maps, while others are trained on point clouds sampled from the ABC dataset. ``A-N'' denotes ABC-NEF dataset and ``A-N-L'' denotes ABC-NEF-Line dataset. }
	\label{tab:comparison}%
	\vspace{-0.5cm}
\end{table}

\subsection{Ablation Studies}
\label{sec:ablation}
We perform ablation studies to verify the inclusion of each loss and design. 
W-MSE loss in Eqn.~\ref{eq:wmse_loss} is essential to learn the NEF due to the imbalanced edge and non-edge pixels (rays). Without W-MSE loss, the training of NEF would suffer from degenerating of predicting all-zero fields. 
Therefore, we take NEF with W-MSE loss as the baseline version, and evaluate sparsity and consistency losses.

For better visualization, we compare the quality of edge densities by illustrating the rendered depth maps. Since a depth map is essentially rendered by accumulated NEF densities along rays, and exactly conveys the spatial distribution of edge density.
As shown in Fig.~\ref{fig:ablation_nerf}, the network may generate random noisy densities in the scene without sparsity regularization. 
Without consistency loss, the network is trained to fit the incorrect ``ground truth'', missing occluded edges, thus overfit 2D edges in each view and fails to reconstruct consistent 3D edges. 

\begin{figure}[t]
	\centering
	\includegraphics[width=0.9\linewidth]{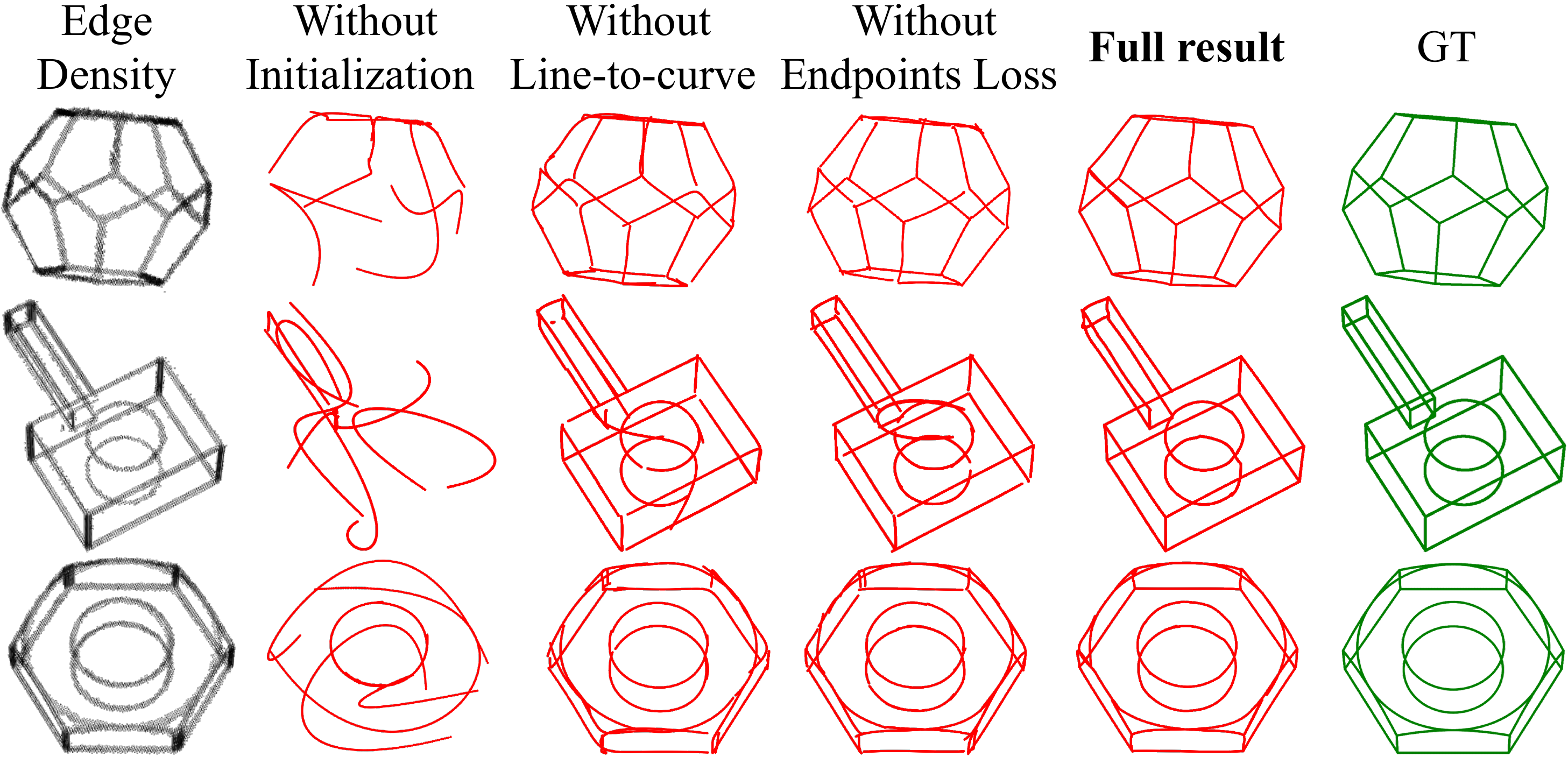}
	\caption{Based on 3D edge points, we show the reconstructed parametric curves of ablations by excluding critical designs from the full version. 
	}
	\label{fig:ablation_curve}
	\vspace{-0.2cm}
\end{figure}

After getting 3D edge points, we reconstruct parametric curves in a coarse-to-fine manner. 
We demonstrate the necessity of our designs by removing each part individually. We show the optimization results with and without the coarse-level initialization, the line-to-curve strategy, and the endpoint loss in Fig.~\ref{fig:ablation_curve}. Quantitative results of the selected samples are shown in Table~\ref{tab:ablation_curve}.
The curves are quite noisy without the coarse-level initialization.  
In the coarse-level, if we try to fit cubic Bézier curves without initializing from straight lines, one cubic Bézier curve may fit multiple connected straight lines. 
Therefore, without the line-to-curve strategy, the total number of curves may be insufficient for global optimization and further influence the endpoints loss. 
The endpoints loss works to refine all curves to be compactly connected. 
With all designed strategies, our full results are clean, compact and fit the geometrical shapes. 

\begin{table}[htbp]
	\centering
	\scalebox{0.95}{
		\begin{tabular}{c|cccc}
			\hline
			Method & CD↓   & F-score↑ & IoU↑ \\
			\hline
			Without Initialization & 0.0734 & 0.5016 & 0.3216 \\
			Without Line-to-curve & 0.0202  & 0.9715 & 0.9387 \\
			Without Endpoints Loss & 0.02   & 0.9805 & 0.959 \\
			Full result & \textbf{0.0189} & \textbf{0.9935} & \textbf{0.9851} \\
			\hline
	\end{tabular}} 
	\caption{Quantitative results of data in Fig.~\ref{fig:ablation_curve}. Initialization improves performance on all metrics significantly. Although the line-to-curve strategy and the endpoints loss seem to bring little improvement, but they help to refine the curves to match the real geometrical shape and to be visually plausible.}
	\label{tab:ablation_curve}
	\vspace{-0.2cm}
\end{table}

\begin{figure}[h!]
	\centering
	\includegraphics[width=0.98\linewidth]{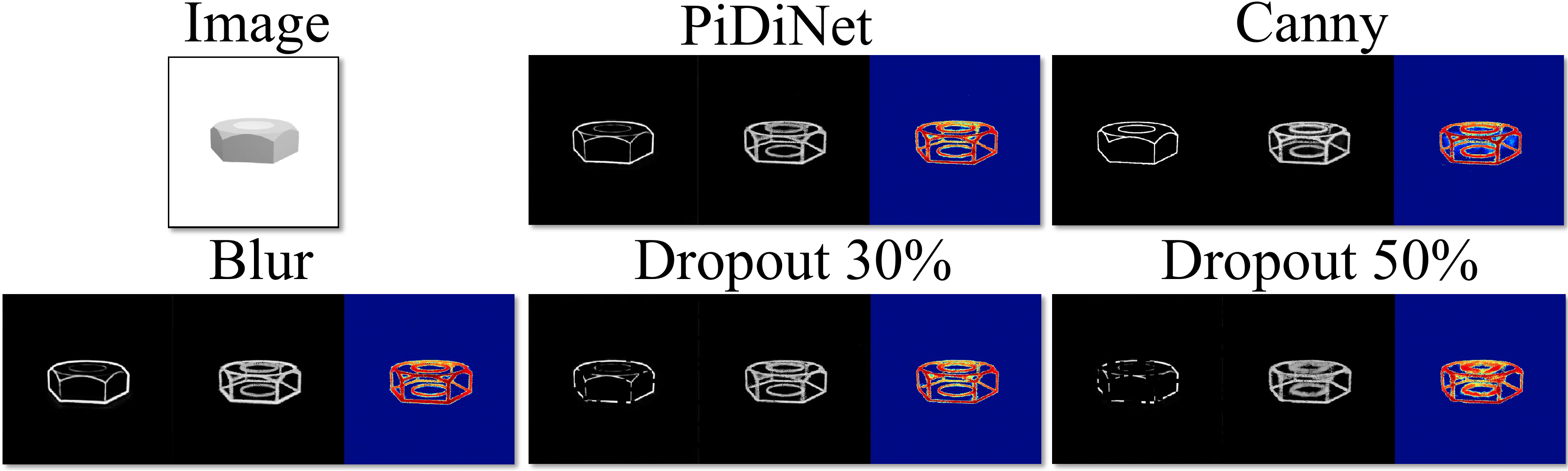}
	\caption{Ablations of Canny detector and low-quality 2D edge maps. For each ablation, from left to right, it shows the detected 2D edge, the rendered edge and depth map (which reveals the distribution of edge densities). 
	}
	\label{fig:edge_detector}
	\vspace{-0.1cm}
\end{figure}

We also conduct ablations on other edge detector (i.e. Canny), noisy 2D edge maps detected on blurred images (Gaussian Blur with a $9 \times 9$ kernel size),  and randomly dropout 30\% and 50\% image (edge) pixels for all views to test the robustness of our method. As in Fig.~\ref{fig:edge_detector}, all alternatives perform reasonably. Even if edge maps are badly broken, it still restores the rough 3D shape. 

\subsection{Real-world Scene}
\begin{figure}[t]
	\centering
	\includegraphics[width=0.95\linewidth]{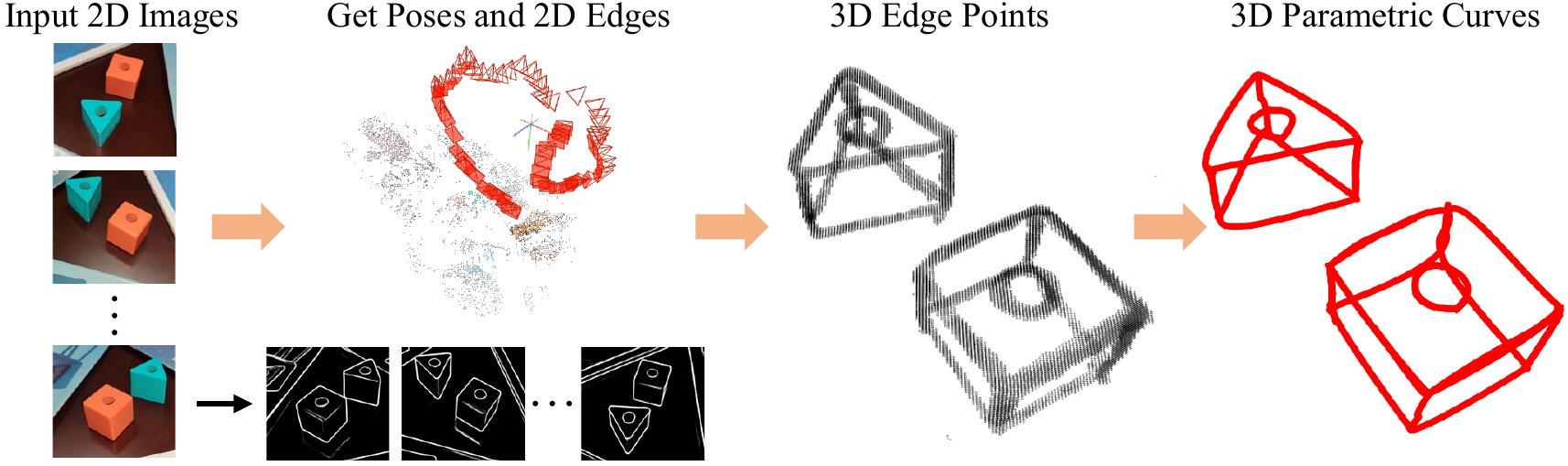}
	\caption{Input a set of multi-view images cut from a video, we use COLMAP~\cite{colmap} to get camera poses, detect 2D edge maps by PiDiNet~\cite{su2021pixel}, and reconstruct 3D curves. 
	}
	\label{fig:real}
	\vspace{-0.2cm}
\end{figure}

We also test the performance of NEF for several collected toys with sharp geometry in the real-world scene. We took a video surrounding and looking at the target toys, and cut about 60 frames as the input. 
We apply COLMAP~\cite{colmap}, a well-known structure-from-motion (SFM) solver, to estimate the camera poses for the input images. We still apply the pre-trained PiDiNeT~\cite{su2021pixel} to extract 2D edge maps, train the NEF, and reconstruct curves from extracted edge points. The process is illustrated in Fig.~\ref{fig:real}. The reconstructed results show the potential of our method to extract 3D edge points and reconstruct parametric curves in real-world scenes, even with camera poses that are not completely correct.

\label{sec:real}
	
	\section{Conclusions}
	We presented the first self-supervised pipeline for 3D parametric curve reconstruction by learning a neural edge field. 
	Self-supervised by only 2D supervisions, our method achieves comparable and even better curve reconstruction than alternatives taking clean and complete point clouds as inputs. Our method shows the potential of generalization ability and leveraging advantages of multi-modal information. 
	The method has limitations in dealing with textured objects, edges inside the objects, and the network architecture would be optimized to be simpler. 
	More discussions are in the Supplementary.
	\vspace{-0.3cm}
	\paragraph{Acknowledgements:} 
	We thank the anonymous reviewers for their valuable comments. 
	This work is supported in part by the National Key Research and Development Program of China (2018AAA0102200), NSFC (62132021, 62002375, 62002376), Natural Science Foundation of Hunan Province of China (2021JJ40696, 2021RC3071, 2022RC1104, 2022RC3061) and NUDT Research Grants (ZK19-30, ZK22-52).
	{\small
		\bibliographystyle{ieee_fullname}
		\bibliography{references}
	}
	\newpage

\twocolumn[
\centering
\Large
\textbf{NEF: Neural Edge Fields for 3D Parametric Curve Reconstruction\\from Multi-view Images} \\
\vspace{0.5em}Supplementary Material \\
\vspace{1.0em}
] %

\begin{appendix}
	
\section{Statistical Analysis for ABC-NEF dataset}
We present more statistics of the contributed ABC-NEF dataset, which consists of 115 distinct and complicated CAD models. Each model can be described by its topology (edges and vertices) as well as the geometry (surfaces and curves). \textit{Edges} are the oriented connections between 2 vertices, with the most to be \textit{sharp edges} where normal changes sharply; \textit{Vertices} are the basic entities, corresponding to points in space. We refer to the original ABC dataset~\cite{koch2019abc} for more detailed explanations.

\begin{figure*}[t]
	\centering
	\includegraphics[width=0.95\linewidth]{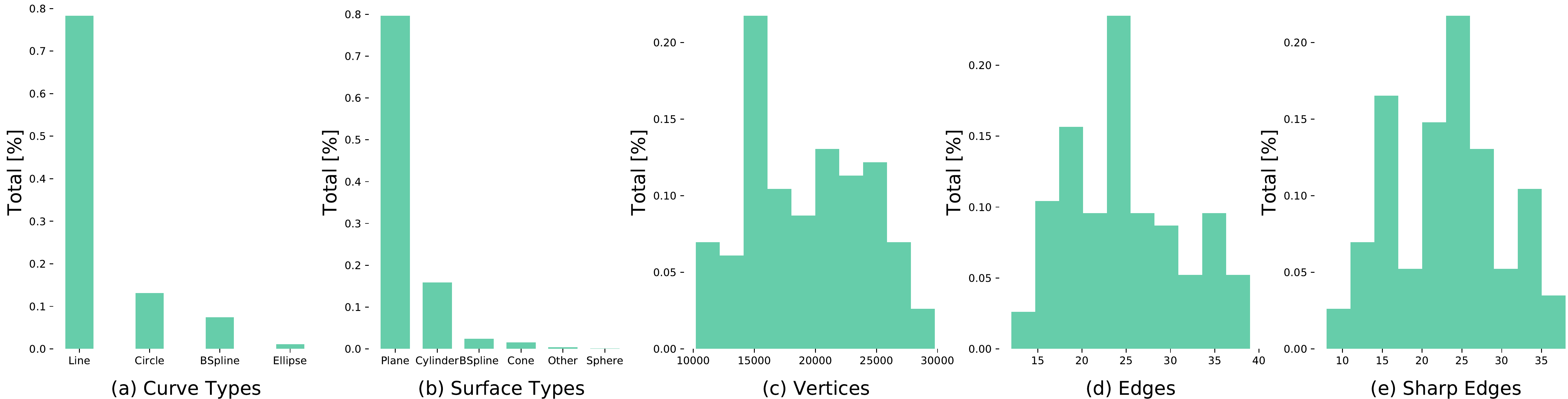}
	\caption{Each model in our dataset is composed of multiple surfaces and feature curves. The first two images show the distribution of types of curves (a) and surfaces (b) over the current ABC-NEF dataset. 
	Histograms over the numbers of vertices (c), edges (d) and sharp edges (e) are presented in last three images. Most edges of the selected models are sharp edges in ABC-NEF dataset, which is qualified as a benchmark of 3D parametric curve reconstruction.}
	\label{fig:statistics}
\end{figure*}

Therefore, we illustrate the distribution of all mentioned attributes in Fig.~\ref{fig:statistics}. The selected models all contain only one part, with medium size of a proper number of vertices $n$ ($10000 <n< 30000$). The major types of edge and surface are line and plane, respectively. We also present the histogram of the vertice, edge and sharp edge numbers in Fig.~\ref{fig:statistics}, to give an impression of the complexity and variety of the dataset. The distribution of the ABC-NEF dataset is close to the original ABC dataset~\cite{koch2019abc}, but as a new benchmark for 3D parametric curve reconstruction, ours focus more on commonly seen objects of medium size with more sharp edges.

\section{Additional Experiments}
Except for this PDF, we also provide several examples for inference using the provided code in the folder ``NEF\_test''. The video demo ``NEF-video-demo.mp4'' also contain 10 examples of the rendered images, detected 2D edge maps, re-rendered 2D edge maps, extracted 3D edge points and reconstructed 3D parametric curves.

Here we provide more training details in Sec.~\ref{sec:details}, experimental results including the ablation study about the required number of views in Sec.~\ref{sec:ablation}, and more comparisons with state-of-the-arts in Sec.~\ref{sec:comparisons}.

\subsection{Training Details}
\label{sec:details}
Our method is implemented in the Pytorch~\cite{paszke2019pytorch} environment and its neural network API PyTorch Lightning~\cite{PyTorch_Lightning}. We sample 1024 rays per batch and train our model for 6 epochs (about 46k iterations) with Adam optimizer~\cite{kingma2014adam} and the learning rate of $5\times10^{-4}$. We use a threshold of $0.7$ to extract point cloud edges from the learned neural edge field with a grid size of $256$. When optimizing all parametric curves, we set $d=4$ in to connect endpoints that are already close enough with a learning rate of $0.5$. All experiments of our method are conducted on a single NVIDIA RTX3080Ti GPU.

\subsection{Ablation Study}
\label{sec:ablation}
\begin{figure*}[t]
	\centering
	\includegraphics[width=0.95\linewidth]{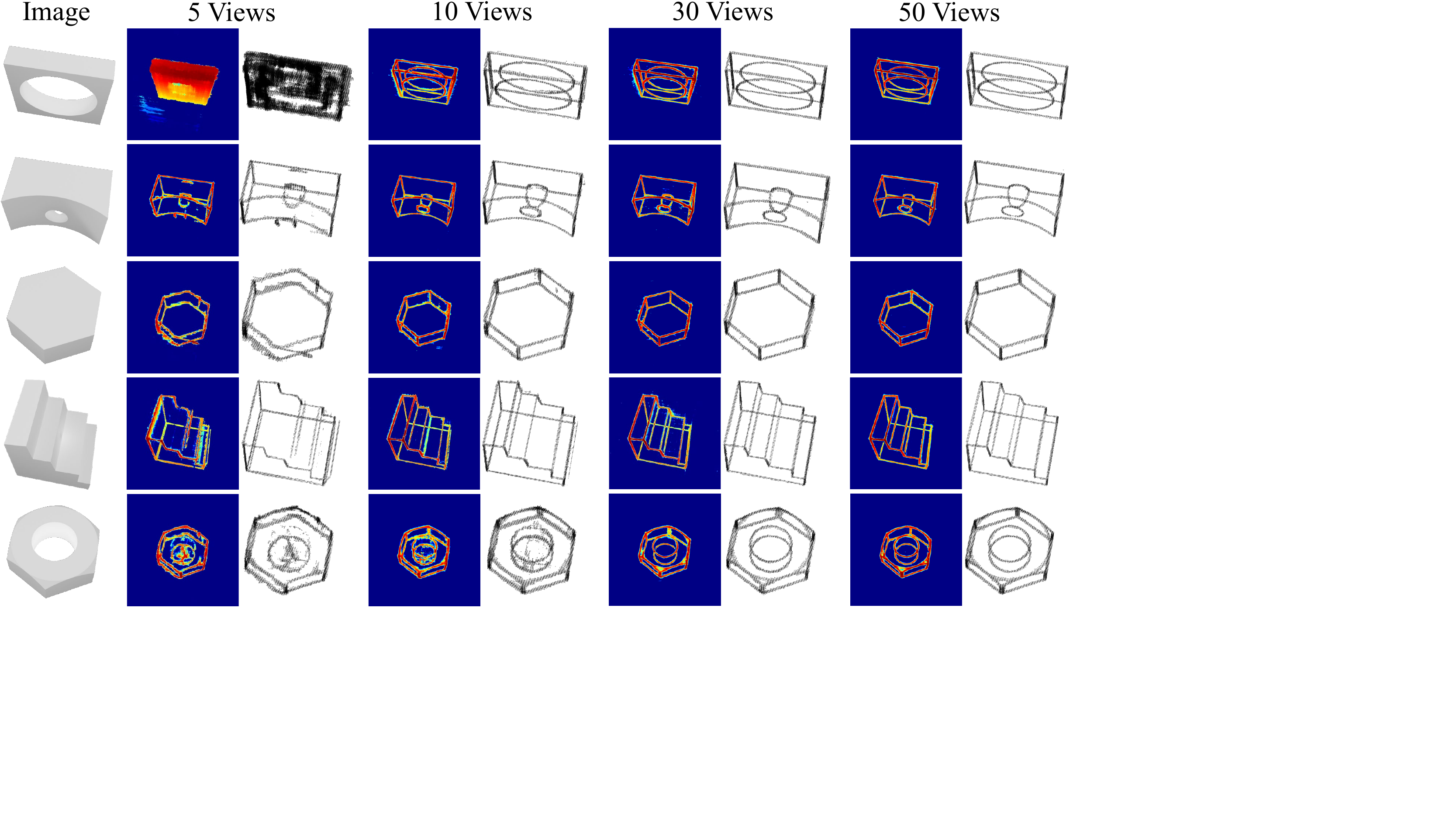}
	\caption{From left to right, we present 2D images in a given view, followed by the rendered depth map and extracted 3D edge points from NEF of 5, 10, 30 and 50 views respectively. Rendered depth maps convey the spatial distribution of the edge density field, and 3D edge points show the extracted geometrical shape. For simple cases, results of 10 views are close to satisfactory, while for complex cases (e.g. the last two rows), more views are required for better performance.}
	\label{fig:ablation_number_view}
	\vspace{-0.2cm}
\end{figure*}

In the proposed ABC-NEF dataset, we sample 50 views for each object by evenly placing cameras on a sphere. Here we conduct an extra ablation study about the required number of views to train neural edge fields (NEF) properly, where the vanilla NeRF~\cite{mildenhall2020nerf} requires about 100 views.
We train the NEF with 5, 10, 30 and 50 views respectively (all evenly distributed) until convergence. As in the main paper, we also observe the spatial distribution of edge density by illustrating rendered depth maps for better visualization. As demonstrated in Fig.~\ref{fig:ablation_number_view}, 5 views are not enough to cover the whole object, and thus cannot get complete and clear edge densities. 10 views can already recover the geometrical shapes for simple cases, but may miss several curves or generate extra noise in objects with relatively complicated shape structures (e.g. the last two rows in Fig.~\ref{fig:ablation_number_view}). 30 views and 50 views are both complete and identical to the real geometrical shape for most cases, which are satisfactory enough. 

Considering that the results of 50 views are slightly clearer, and the time consumption is close for training NEF by 30 views and 50 views until convergence, we finally decide to sample a unified number of 50 views for all cases for better performance, although 10$-$30 views are enough for most simpler cases.  

\subsection{More Comparisons}\label{sec:comparisons}
\begin{figure*}[!h]
	
	\begin{center}
		\begin{minipage}{0.137\linewidth}
			\centerline{Image}
			\vspace{2pt}
			\centerline{\includegraphics[width=\textwidth]{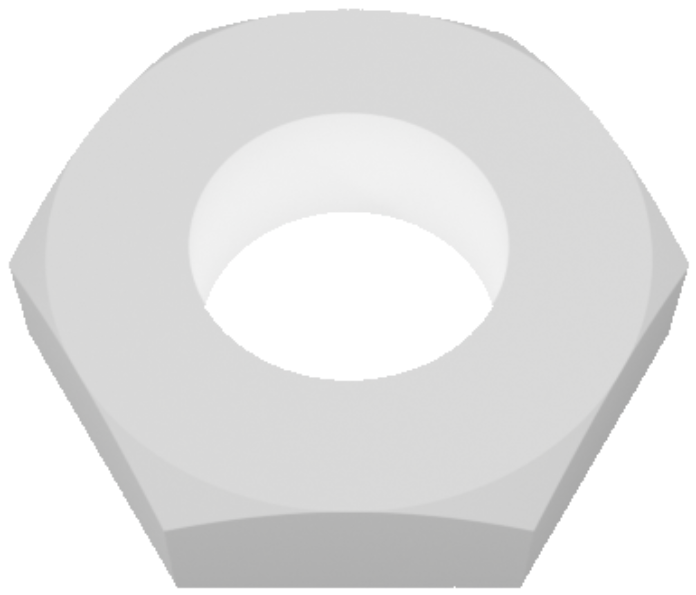}}
			\vspace{1.5pt}
			\centerline{\includegraphics[width=\textwidth]{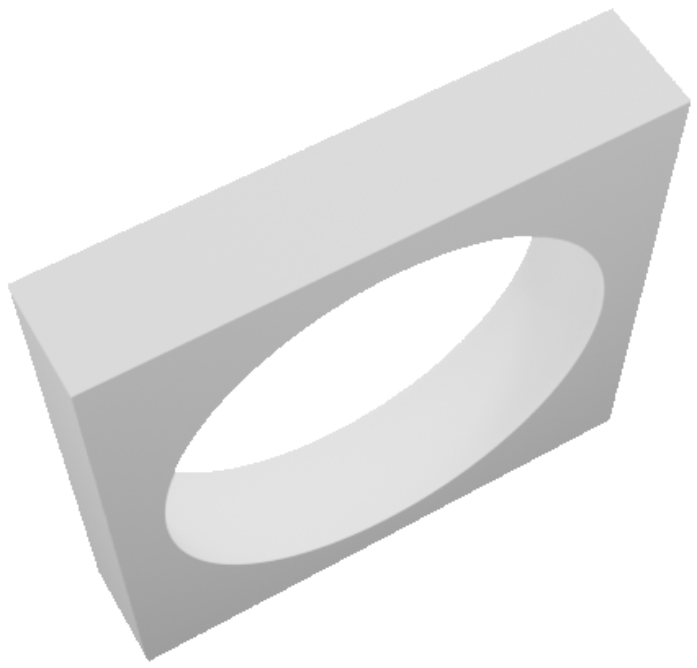}}
			\vspace{1.5pt}
			\centerline{\includegraphics[width=\textwidth]{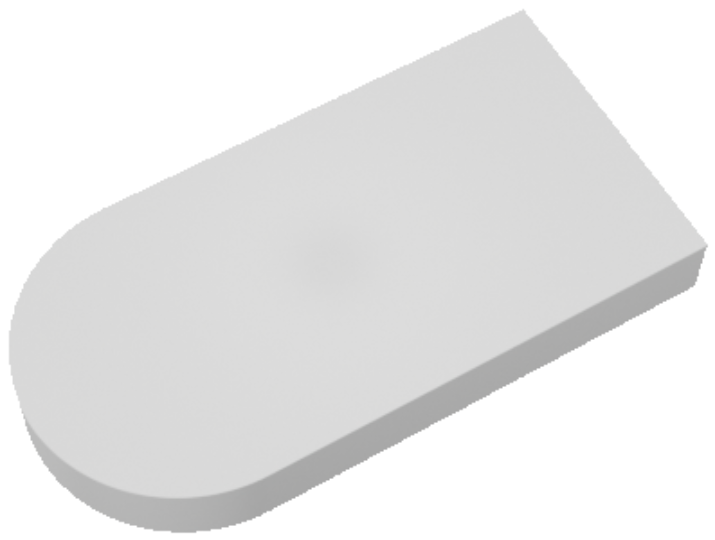}}
			\vspace{1.5pt}
			\centerline{\includegraphics[width=\textwidth]{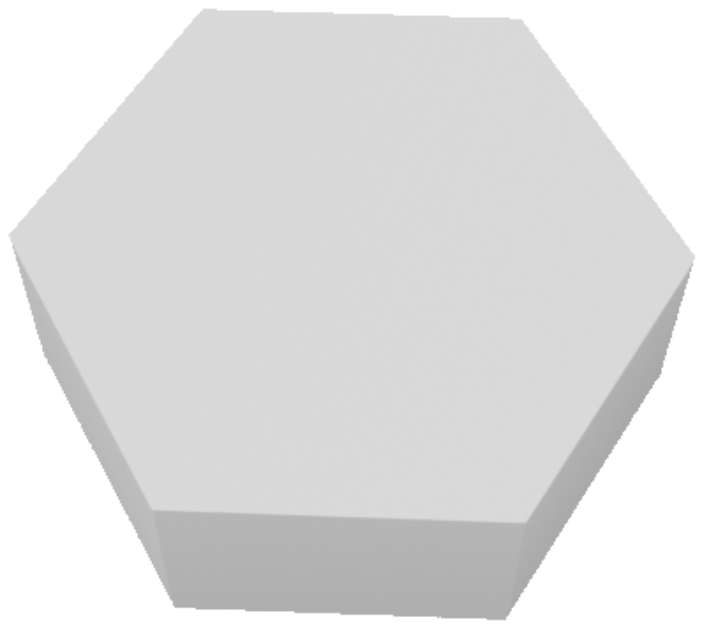}}
			\vspace{1.5pt}
			\centerline{\includegraphics[width=\textwidth]{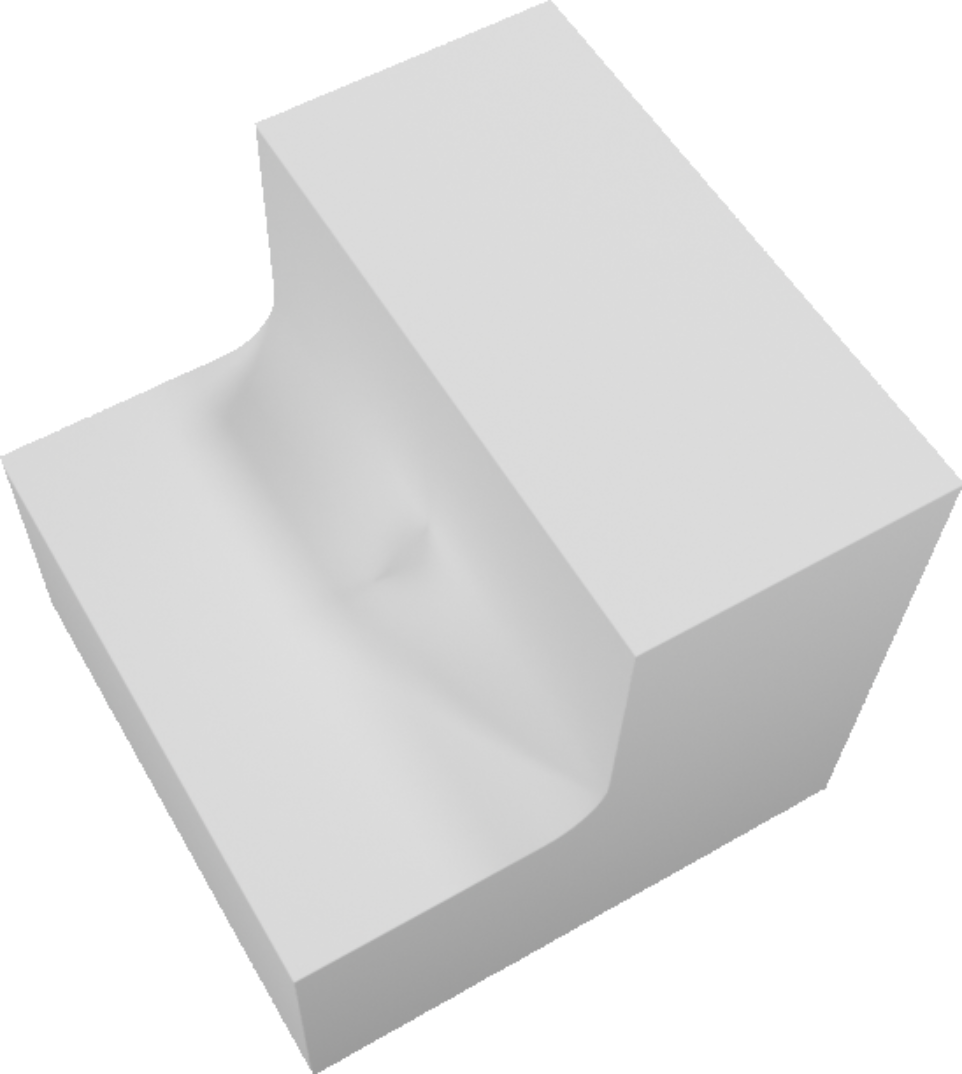}}
			\vspace{1.5pt}
			\centerline{\includegraphics[width=\textwidth]{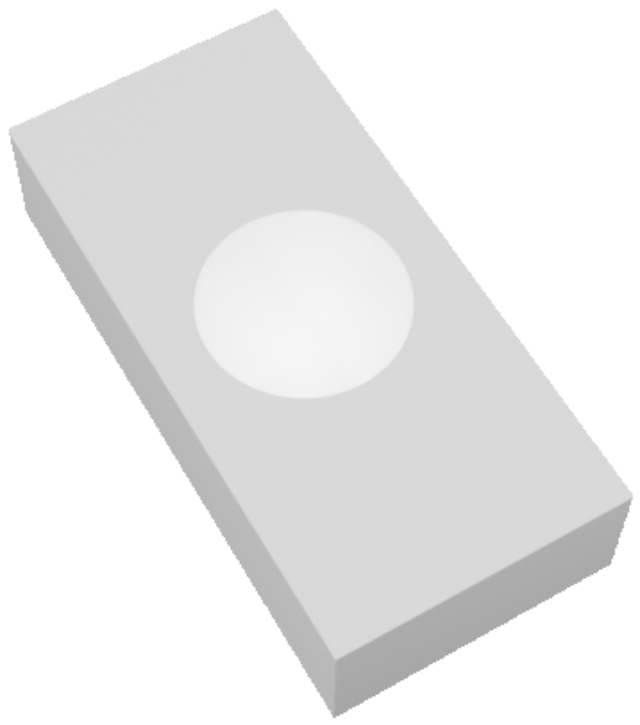}}
			\vspace{1.5pt}
			\centerline{\includegraphics[width=\textwidth]{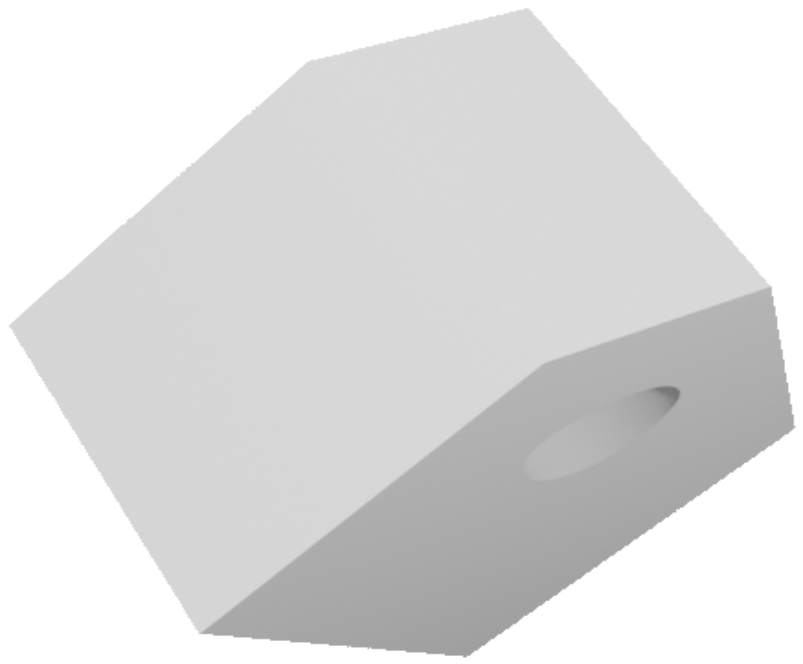}}
			\vspace{1.5pt}
			\centerline{\includegraphics[width=\textwidth]{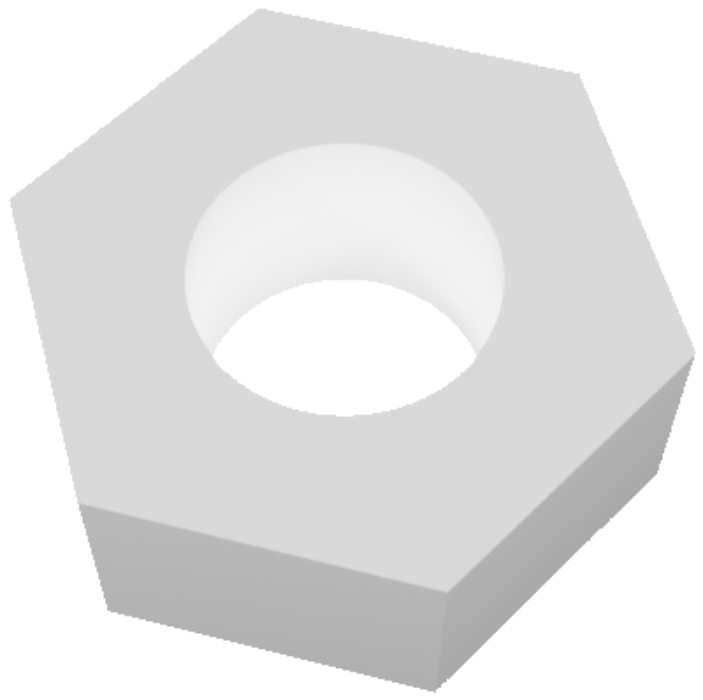}}
			\vspace{1.5pt}
			\centerline{\includegraphics[width=\textwidth]{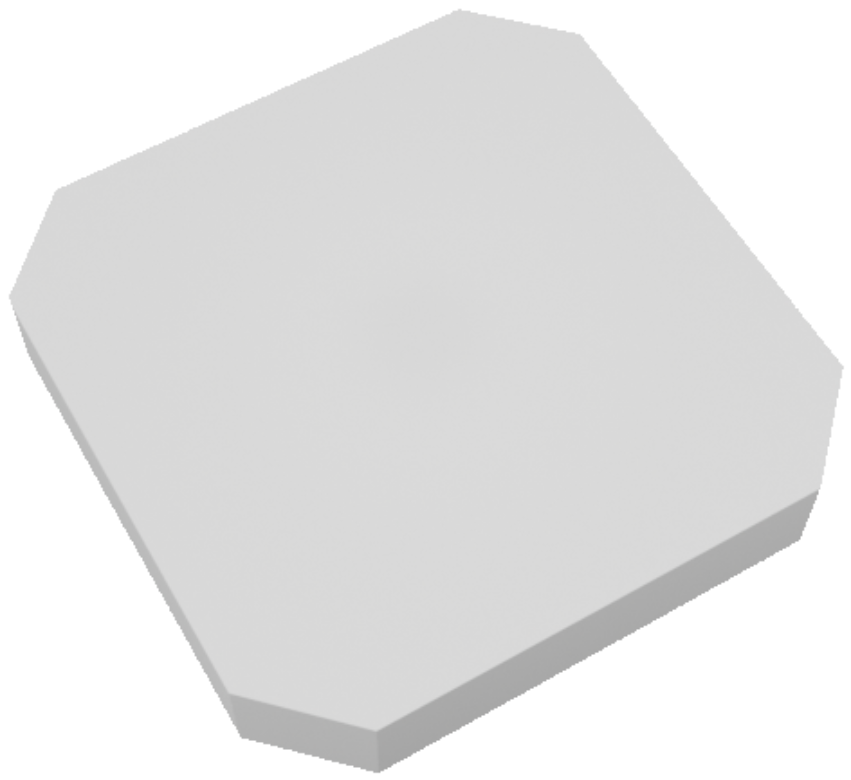}}
		\end{minipage}
		\begin{minipage}{0.137\linewidth}
			\centerline{PIE-NET}
			\vspace{2pt}
			\centerline{\includegraphics[width=\textwidth]{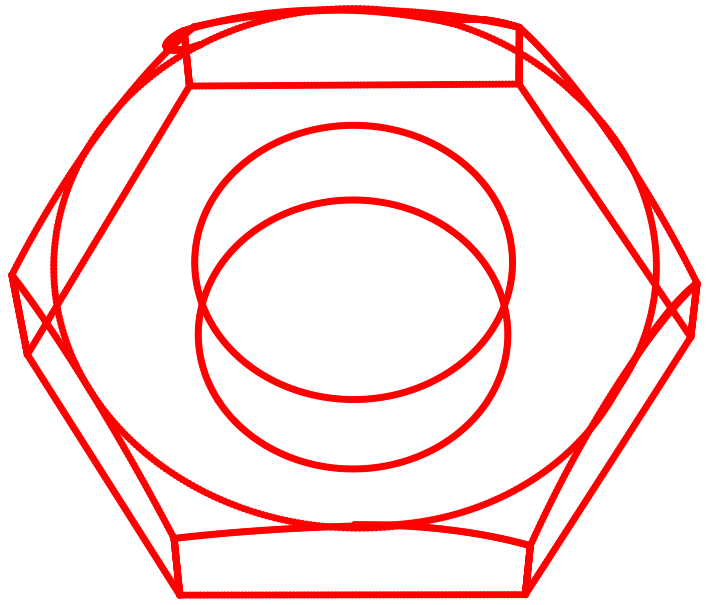}}
			\vspace{1.5pt}
			\centerline{\includegraphics[width=\textwidth]{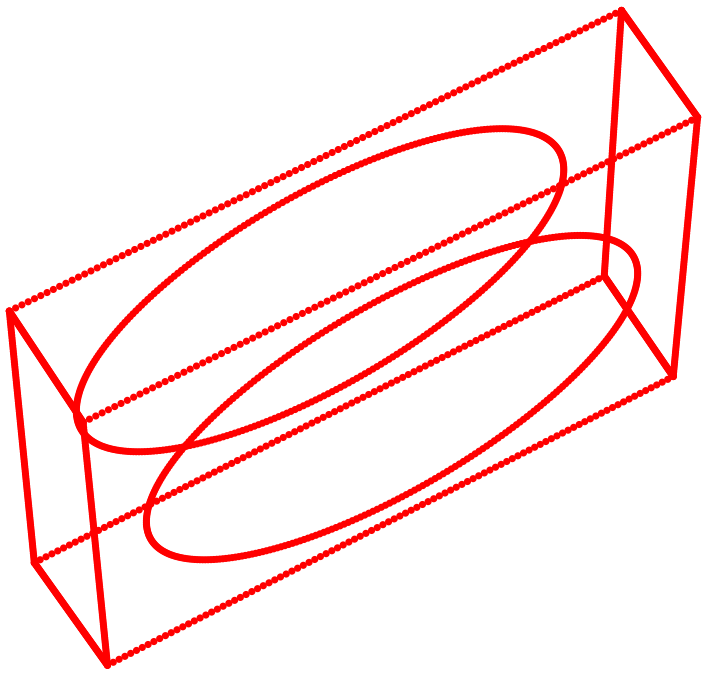}}
			\vspace{1.5pt}
			\centerline{\includegraphics[width=\textwidth]{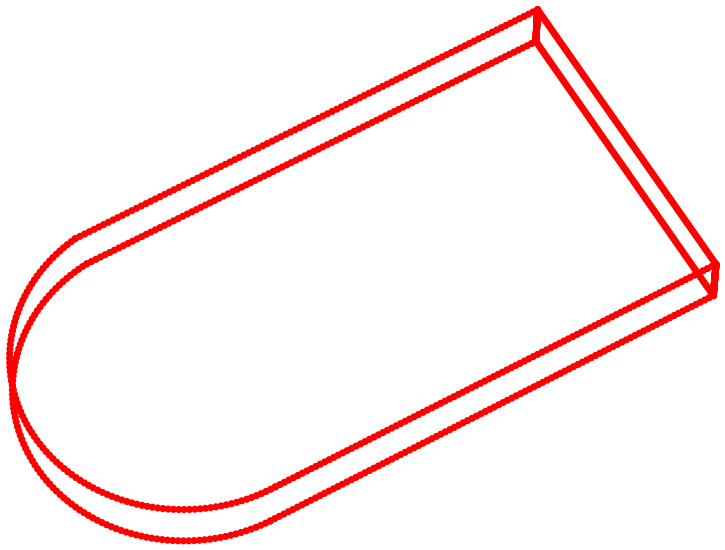}}
			\vspace{1.5pt}
			\centerline{\includegraphics[width=\textwidth]{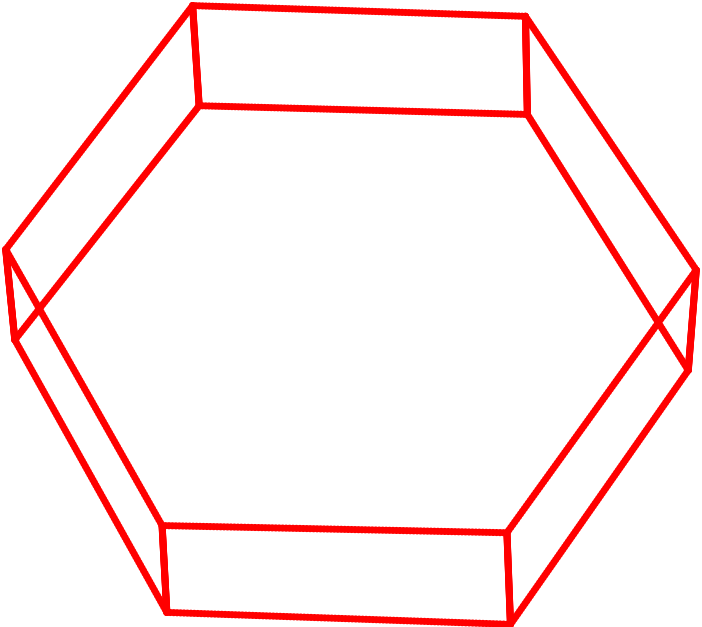}}
			\vspace{1.5pt}
			\centerline{\includegraphics[width=\textwidth]{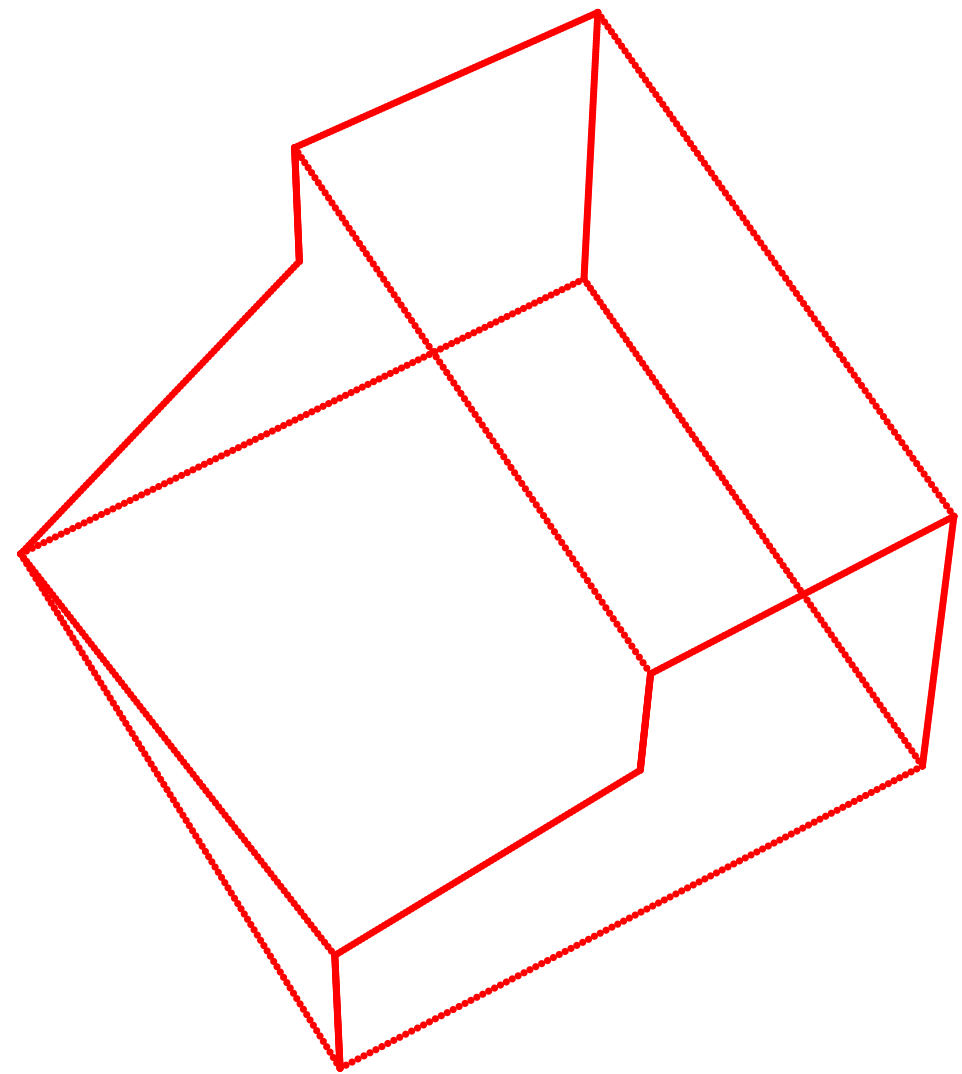}}
			\vspace{1.5pt}
			\centerline{\includegraphics[width=\textwidth]{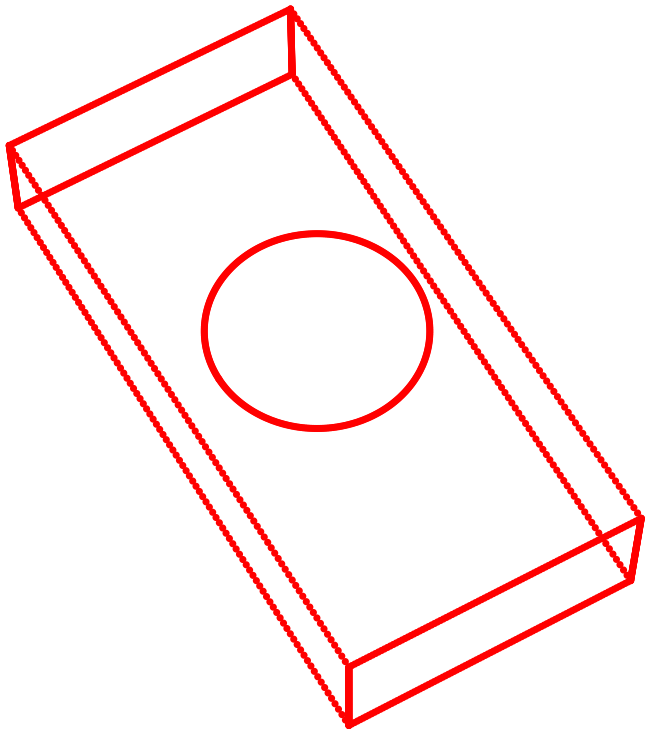}}
			\vspace{1.5pt}
			\centerline{\includegraphics[width=\textwidth]{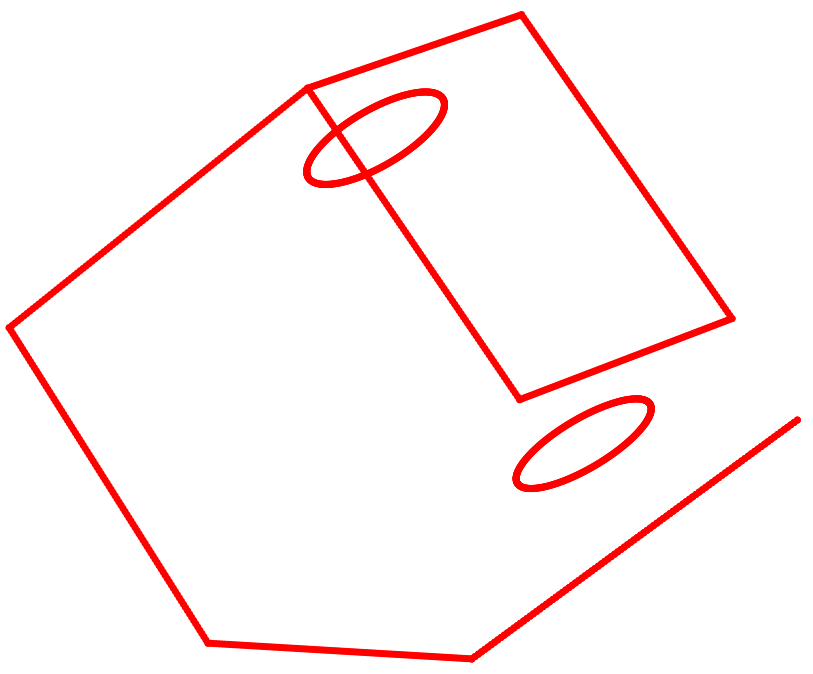}}
			\vspace{1.5pt}
			\centerline{\includegraphics[width=\textwidth]{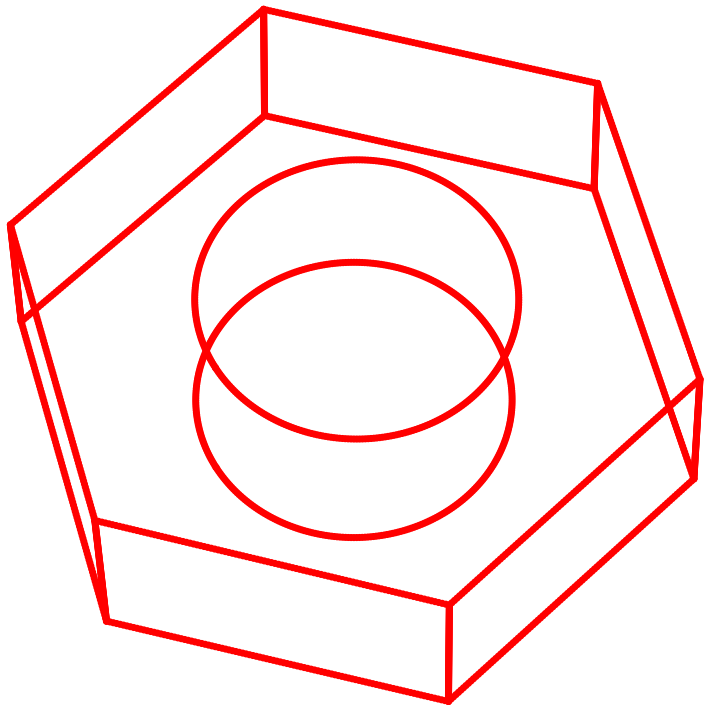}}
			\vspace{1.5pt}
			\centerline{\includegraphics[width=\textwidth]{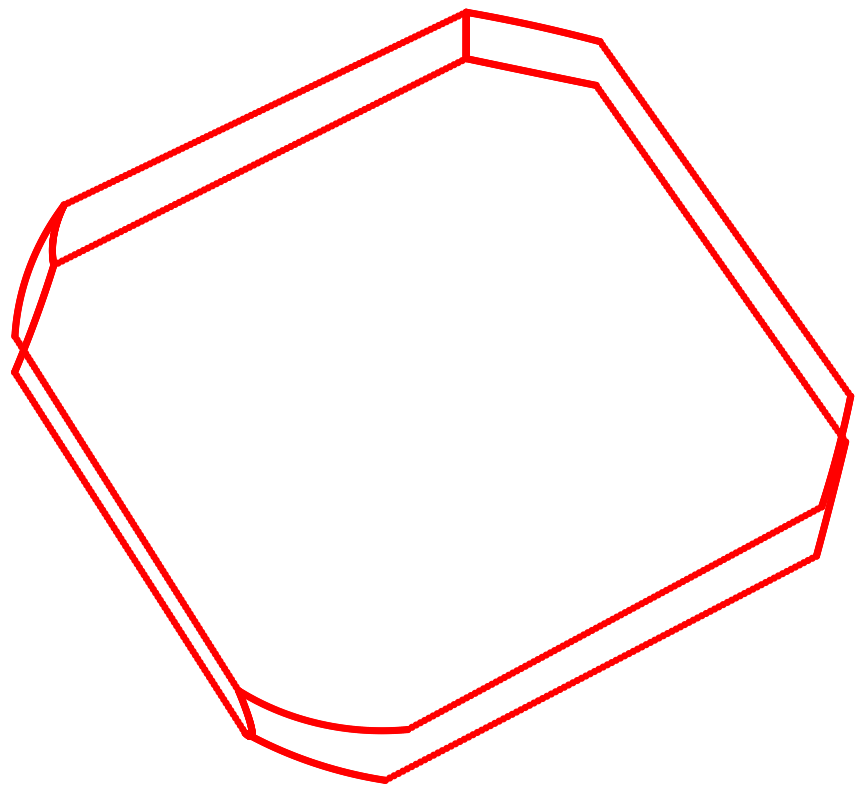}}
			
		\end{minipage}
		\begin{minipage}{0.137\linewidth}
			\centerline{PC2WF}
			\vspace{2pt}
			\centerline{\includegraphics[width=\textwidth]{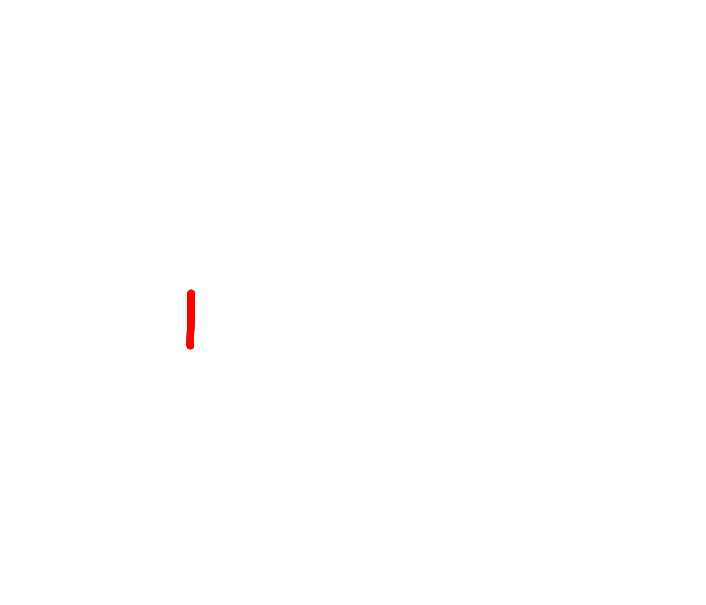}}
			\vspace{1.5pt}
			\centerline{\includegraphics[width=\textwidth]{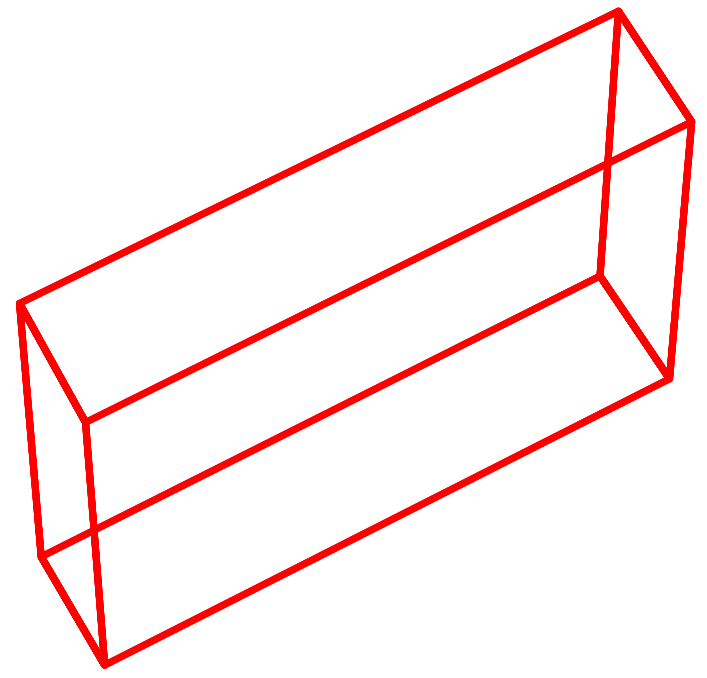}}
			\vspace{1.5pt}
			\centerline{\includegraphics[width=\textwidth]{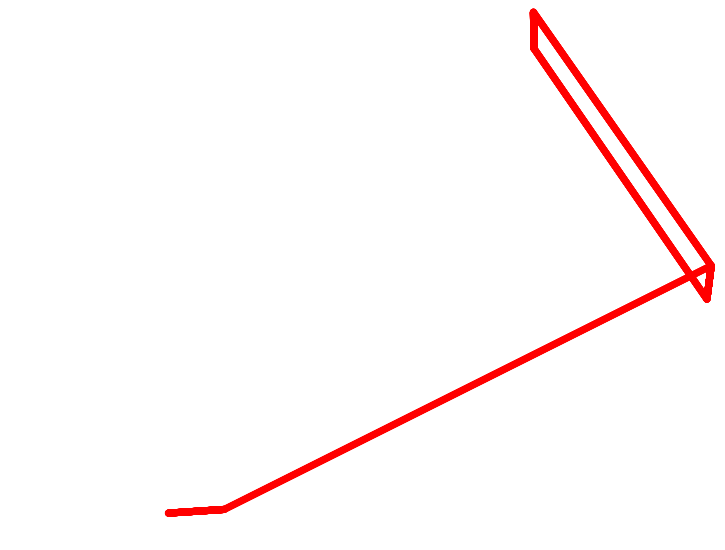}}
			\vspace{1.5pt}
			\centerline{\includegraphics[width=\textwidth]{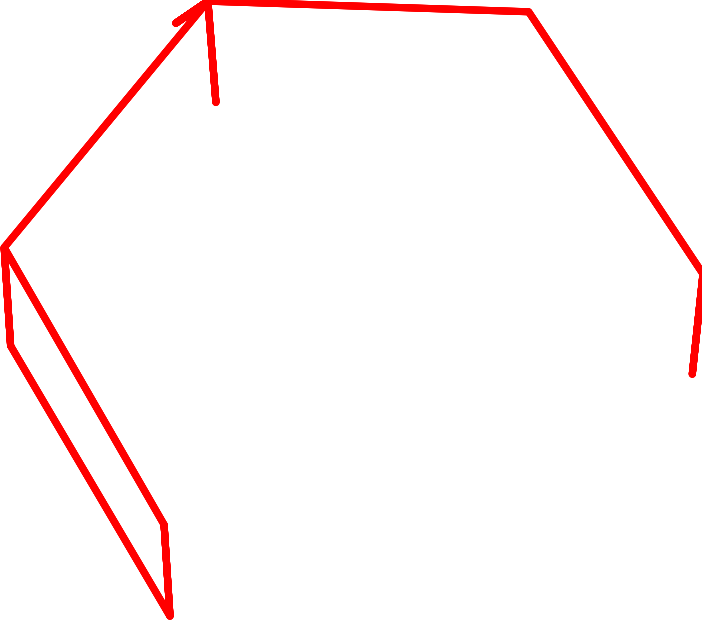}}
			\vspace{1.5pt}
			\centerline{\includegraphics[width=\textwidth]{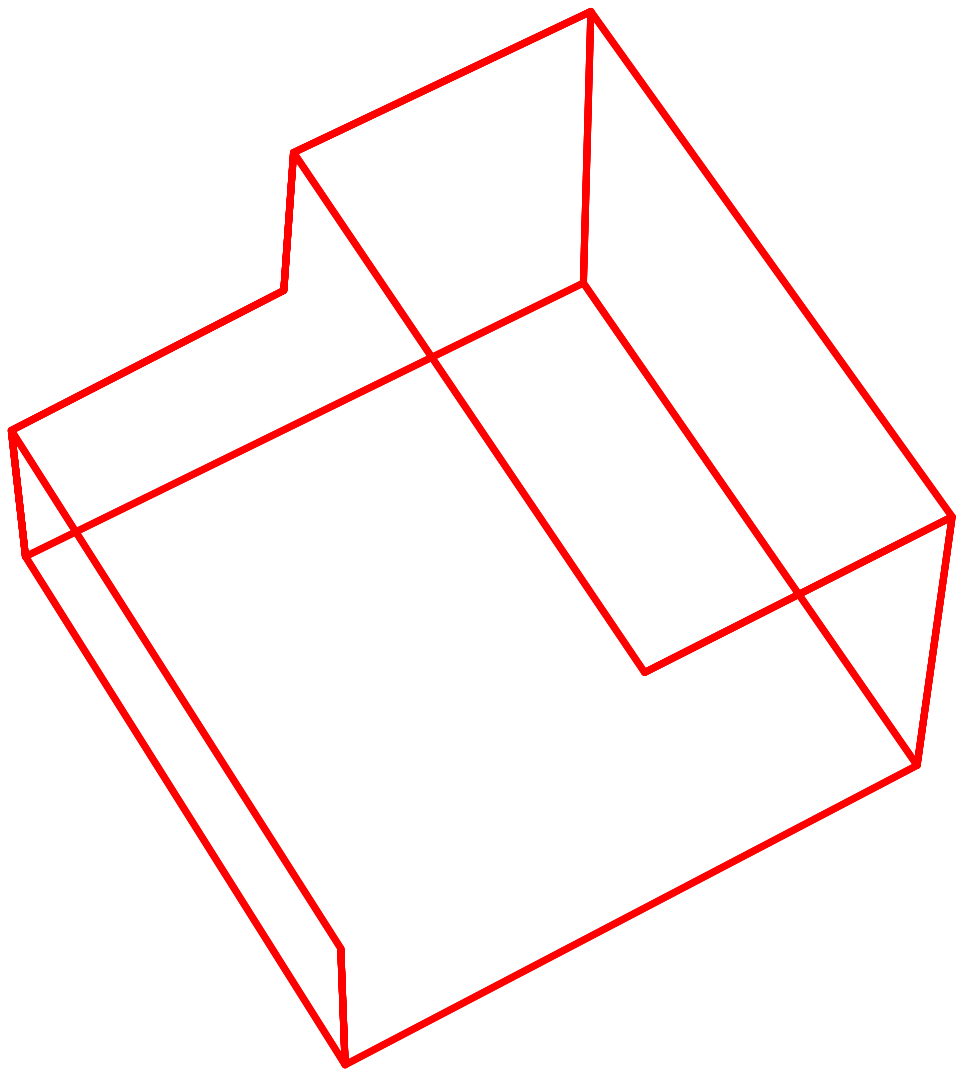}}
			\vspace{1.5pt}
			\centerline{\includegraphics[width=\textwidth]{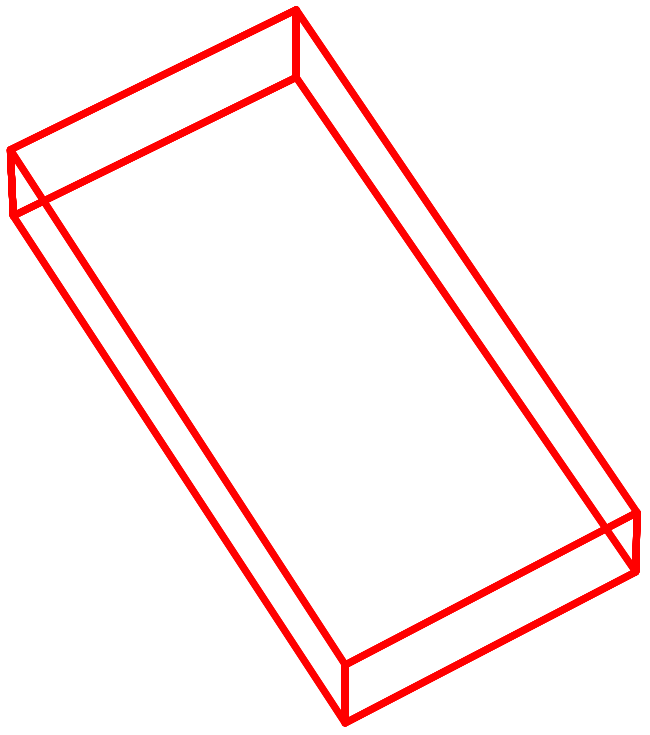}}
			\vspace{1.5pt}
			\centerline{\includegraphics[width=\textwidth]{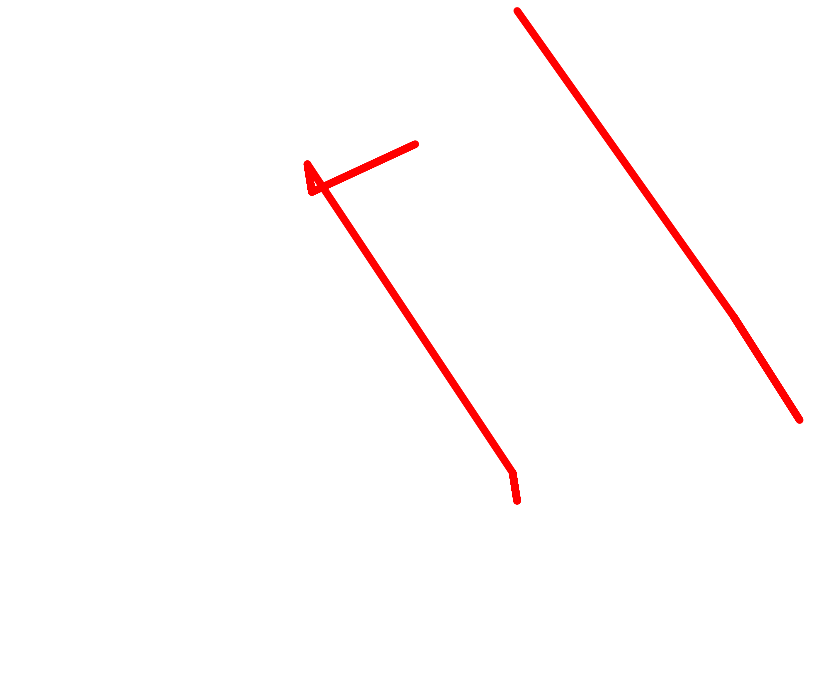}}
			\vspace{1.5pt}
			\centerline{\includegraphics[width=\textwidth]{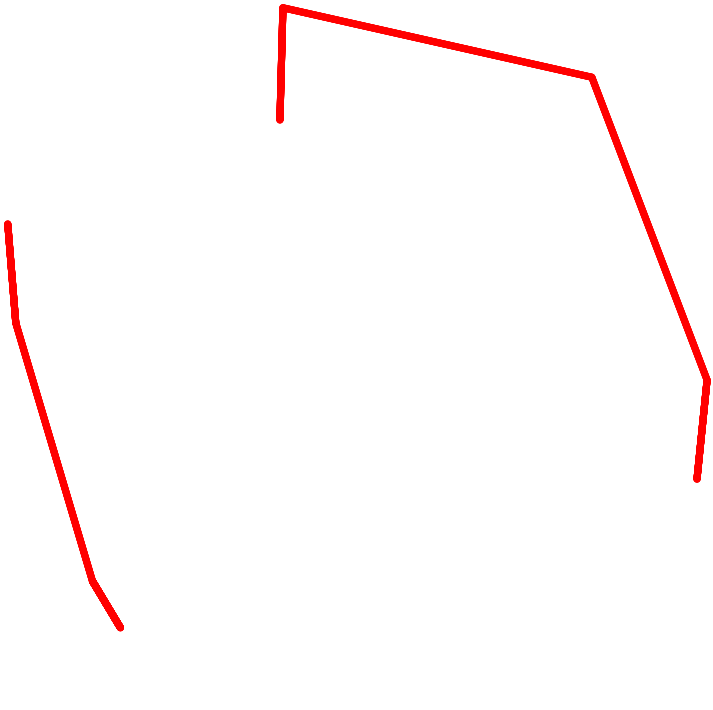}}
			\vspace{1.5pt}
			\centerline{\includegraphics[width=\textwidth]{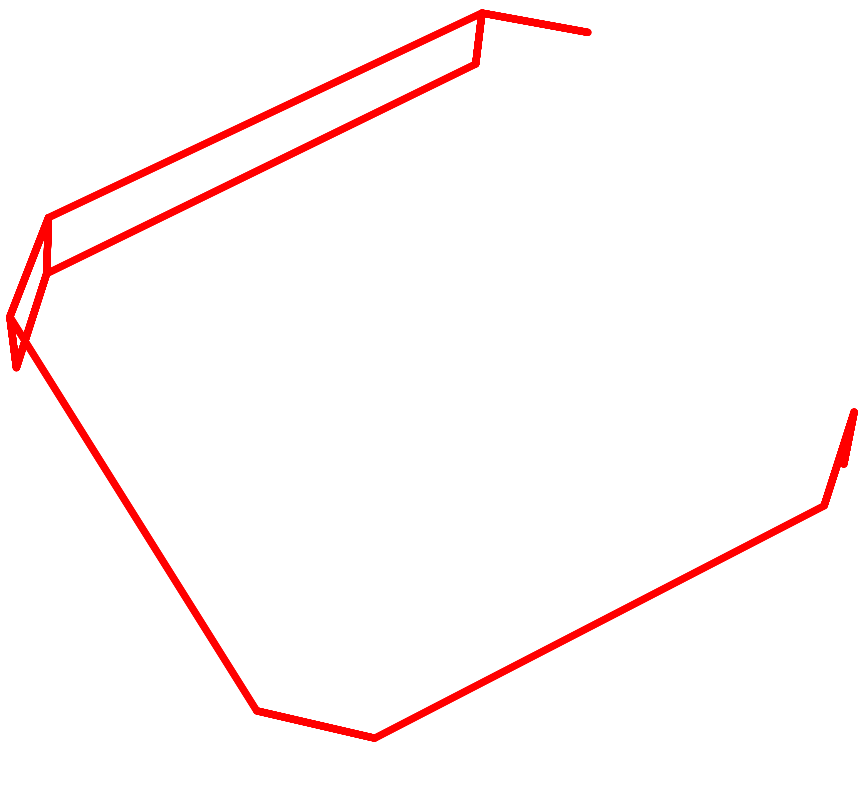}}
			
		\end{minipage}
		\begin{minipage}{0.137\linewidth}
			\centerline{DEF}
			\vspace{2pt}
			\centerline{\includegraphics[width=\textwidth]{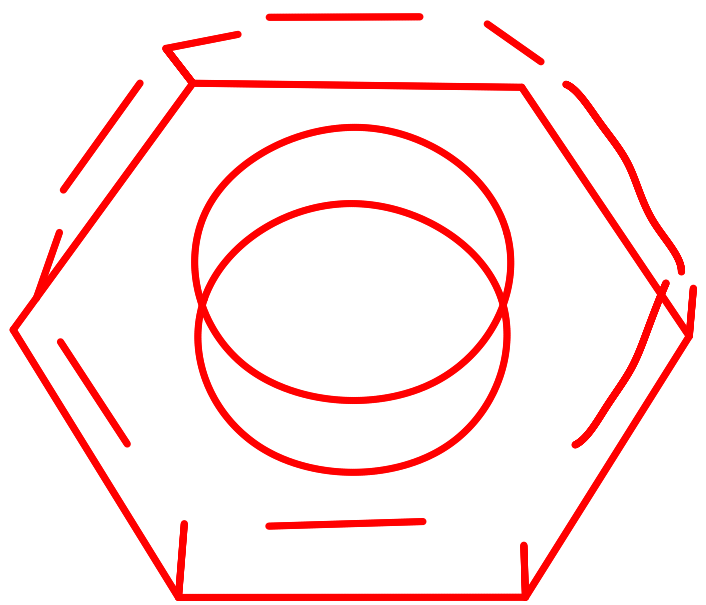}}
			\vspace{1.5pt}
			\centerline{\includegraphics[width=\textwidth]{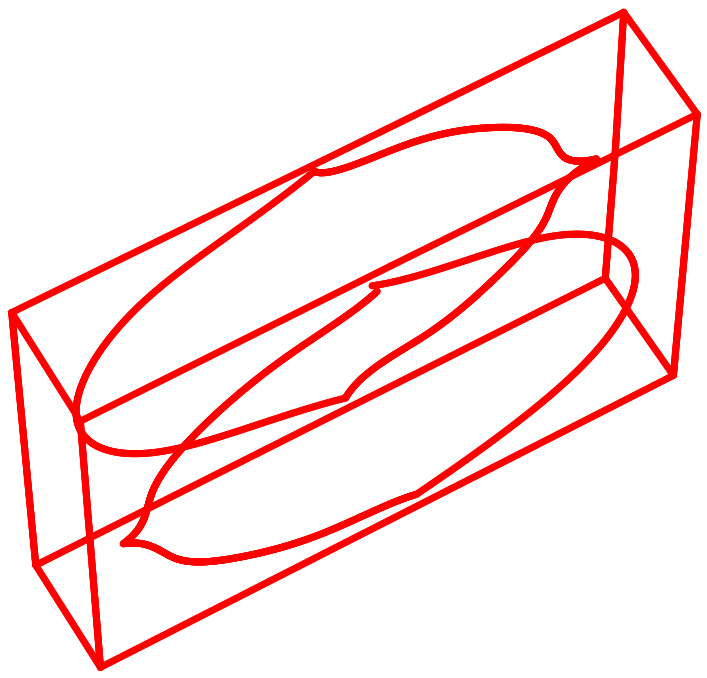}}
			\vspace{1.5pt}
			\centerline{\includegraphics[width=\textwidth]{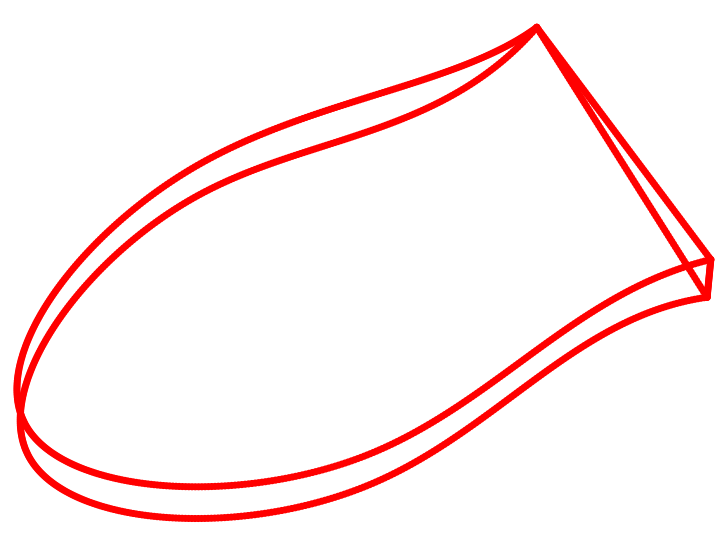}}
			\vspace{1.5pt}
			\centerline{\includegraphics[width=\textwidth]{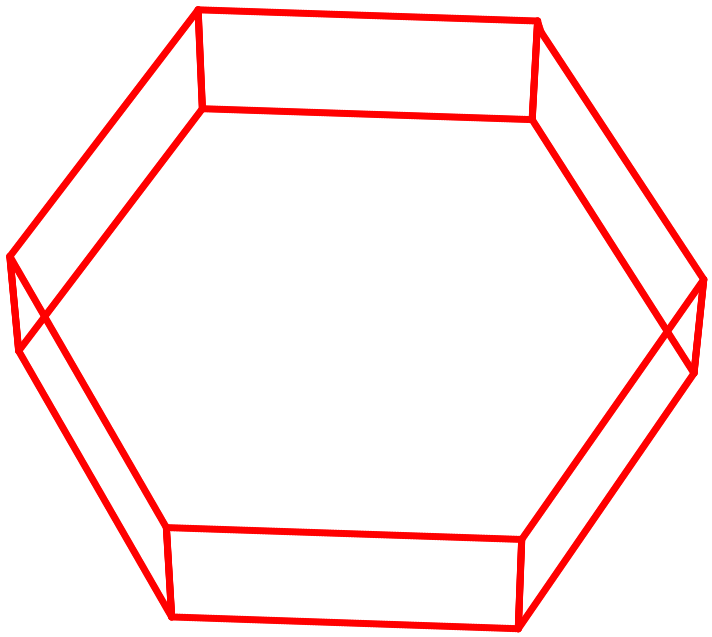}}
			\vspace{1.5pt}
			\centerline{\includegraphics[width=\textwidth]{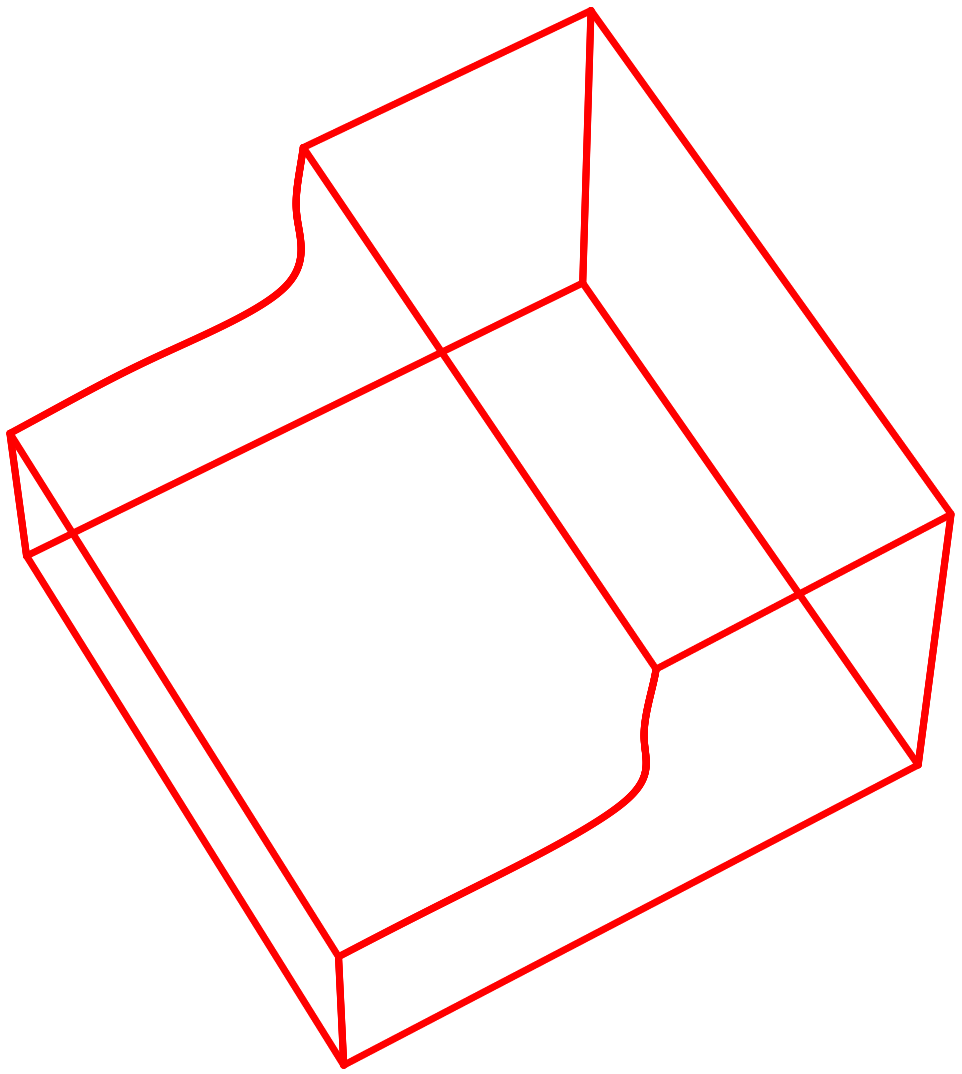}}
			\vspace{1.5pt}
			\centerline{\includegraphics[width=\textwidth]{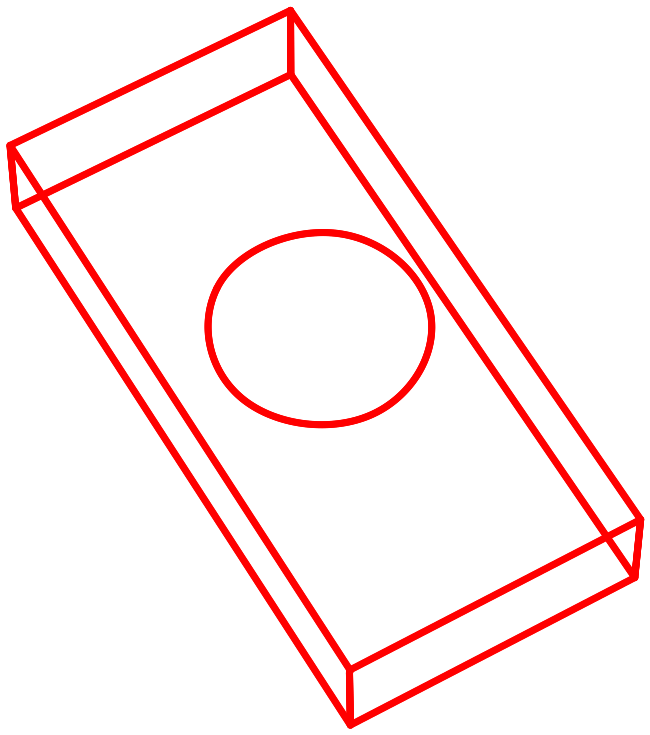}}
			\vspace{1.5pt}
			\centerline{\includegraphics[width=\textwidth]{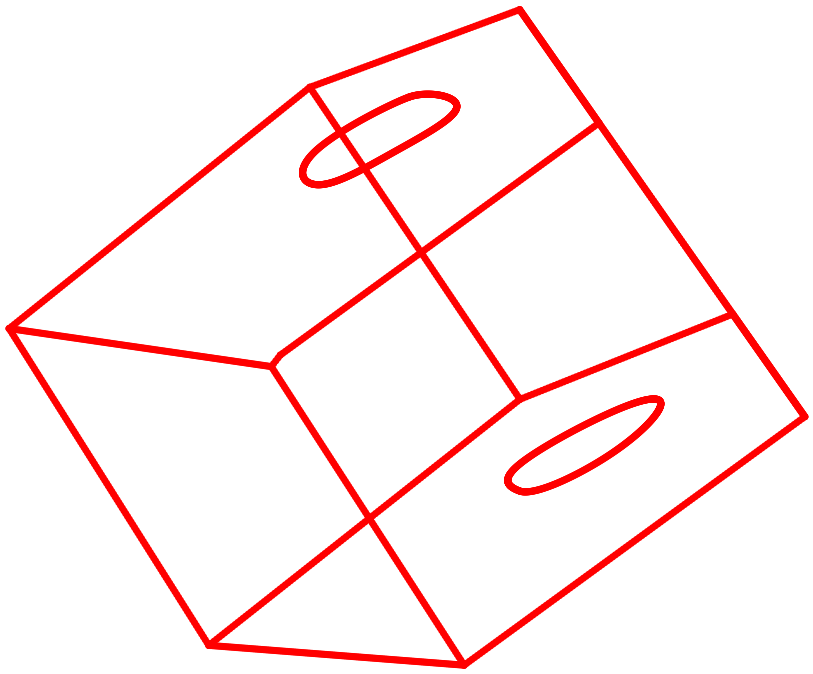}}
			\vspace{1.5pt}
			\centerline{\includegraphics[width=\textwidth]{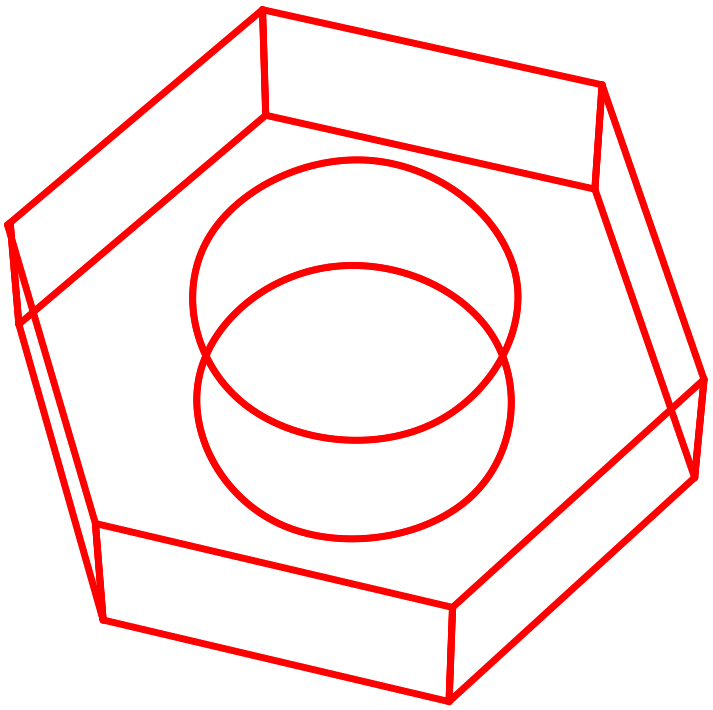}}
			\vspace{1.5pt}
			\centerline{\includegraphics[width=\textwidth]{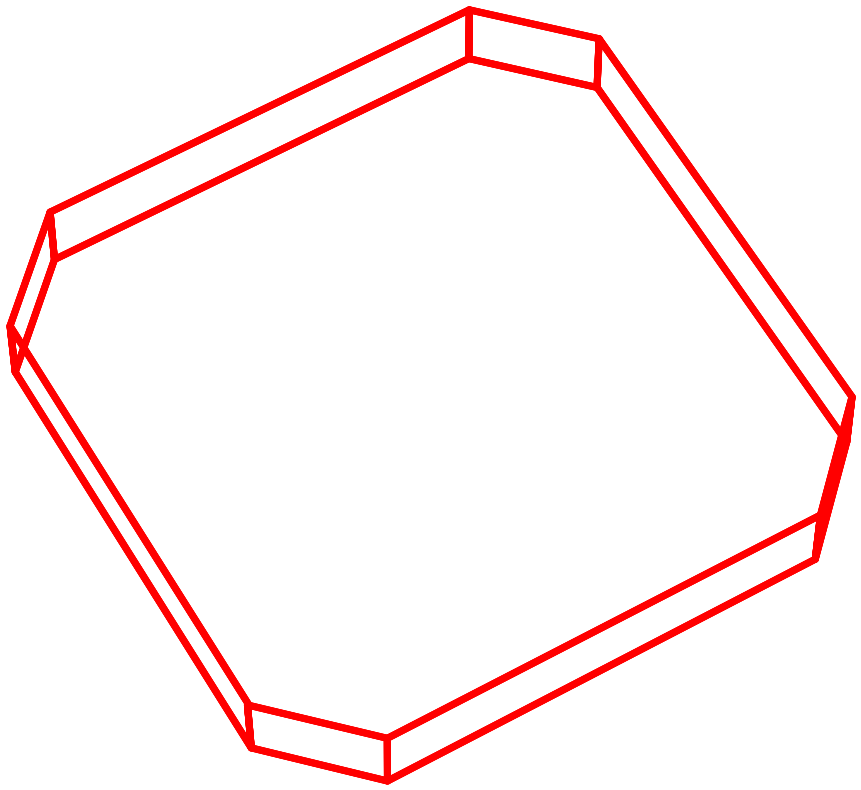}}
			
		\end{minipage}
		\begin{minipage}{0.137\linewidth}
			\centerline{Ours (Curve)}
			\vspace{2pt}
			\centerline{\includegraphics[width=\textwidth]{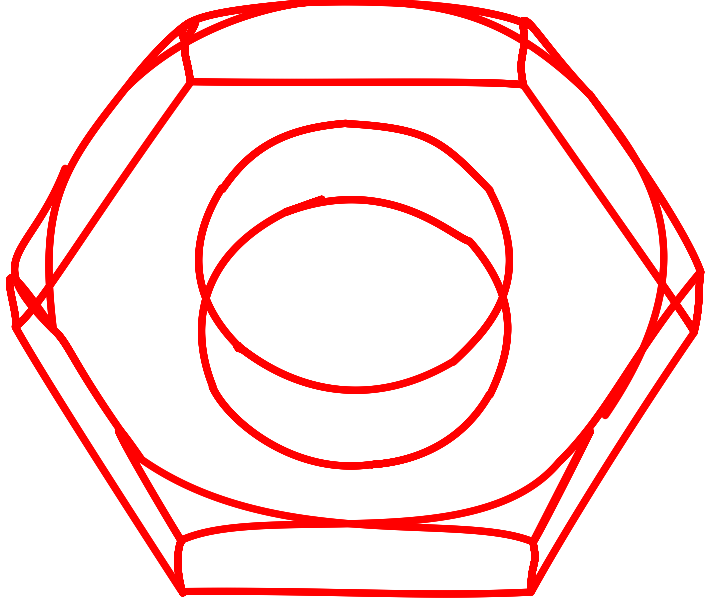}}
			\vspace{1.5pt}
			\centerline{\includegraphics[width=\textwidth]{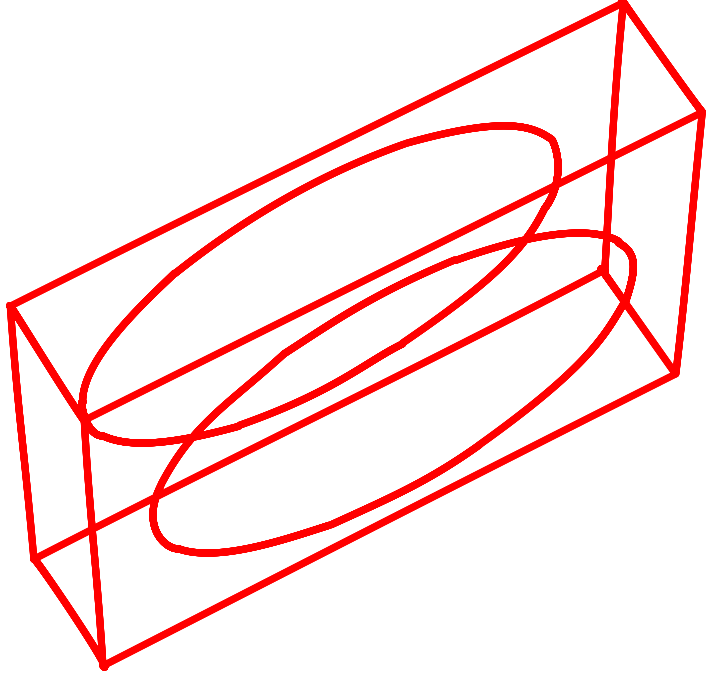}}
			\vspace{1.5pt}
			\centerline{\includegraphics[width=\textwidth]{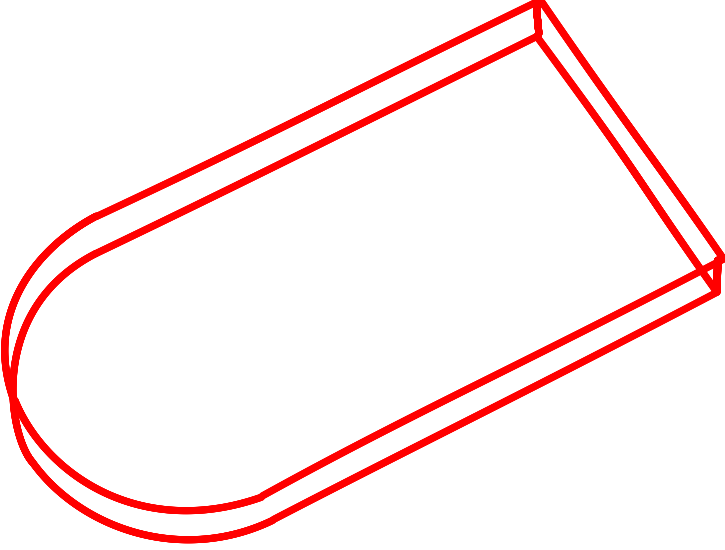}}
			\vspace{1.5pt}
			\centerline{\includegraphics[width=\textwidth]{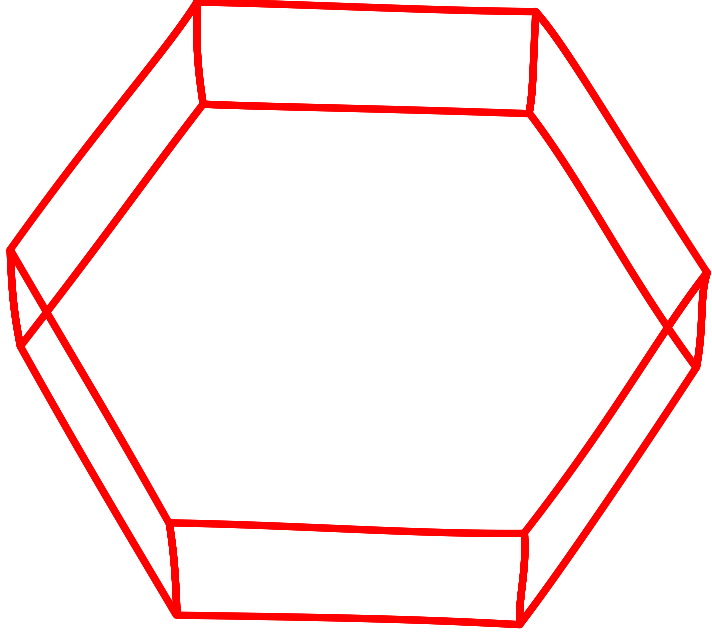}}
			\vspace{1.5pt}
			\centerline{\includegraphics[width=\textwidth]{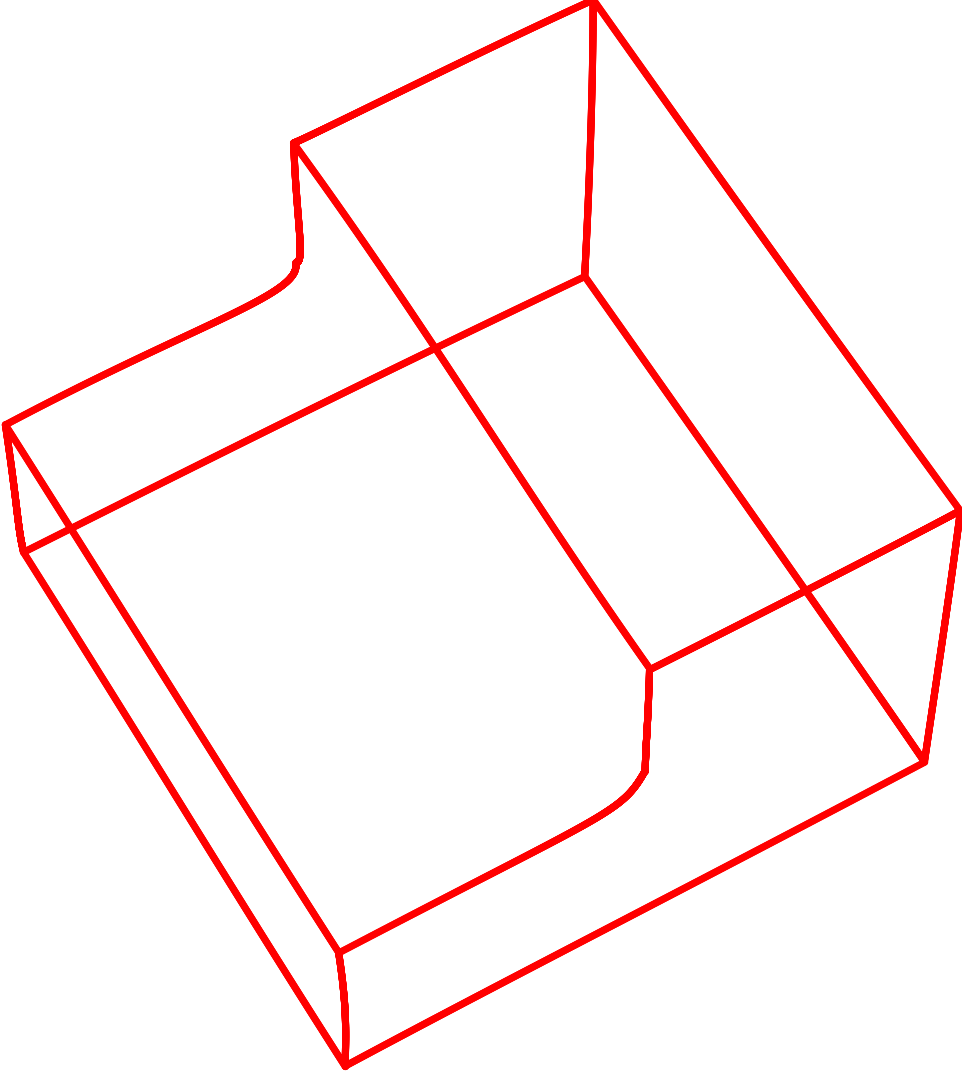}}
			\vspace{1.5pt}
			\centerline{\includegraphics[width=\textwidth]{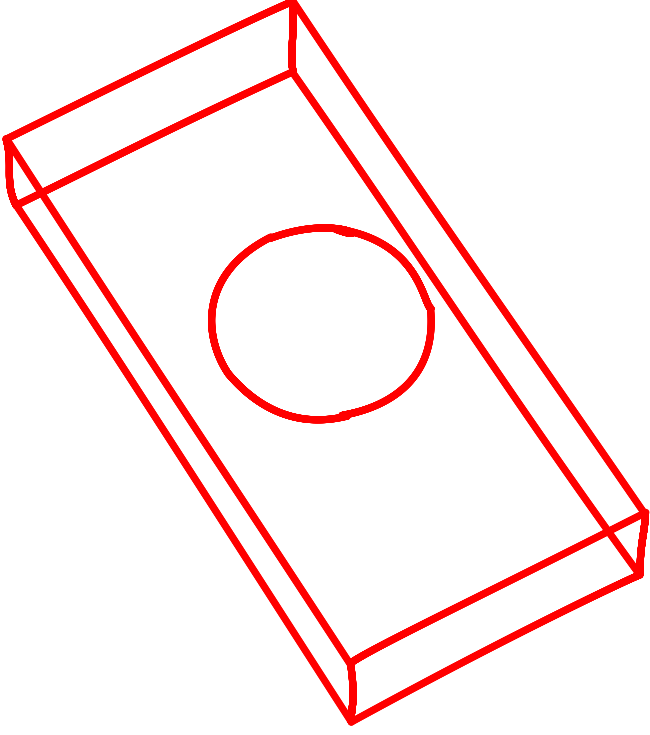}}
			\vspace{1.5pt}
			\centerline{\includegraphics[width=\textwidth]{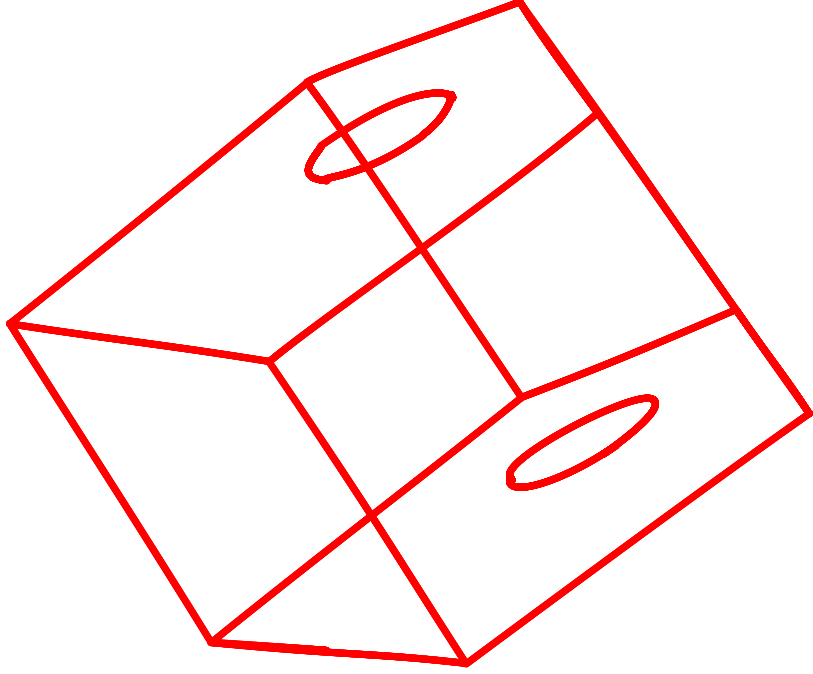}}
			\vspace{1.5pt}
			\centerline{\includegraphics[width=\textwidth]{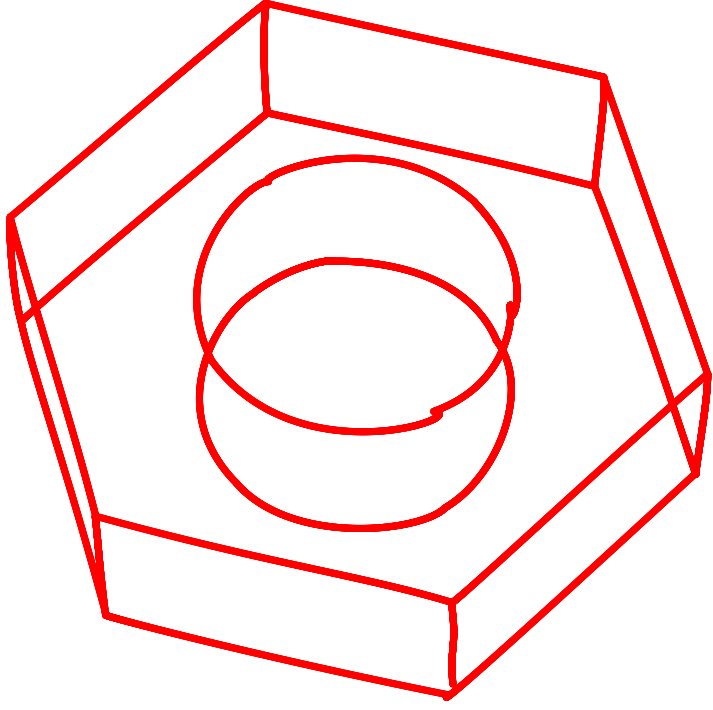}}
			\vspace{1.5pt}
			\centerline{\includegraphics[width=\textwidth]{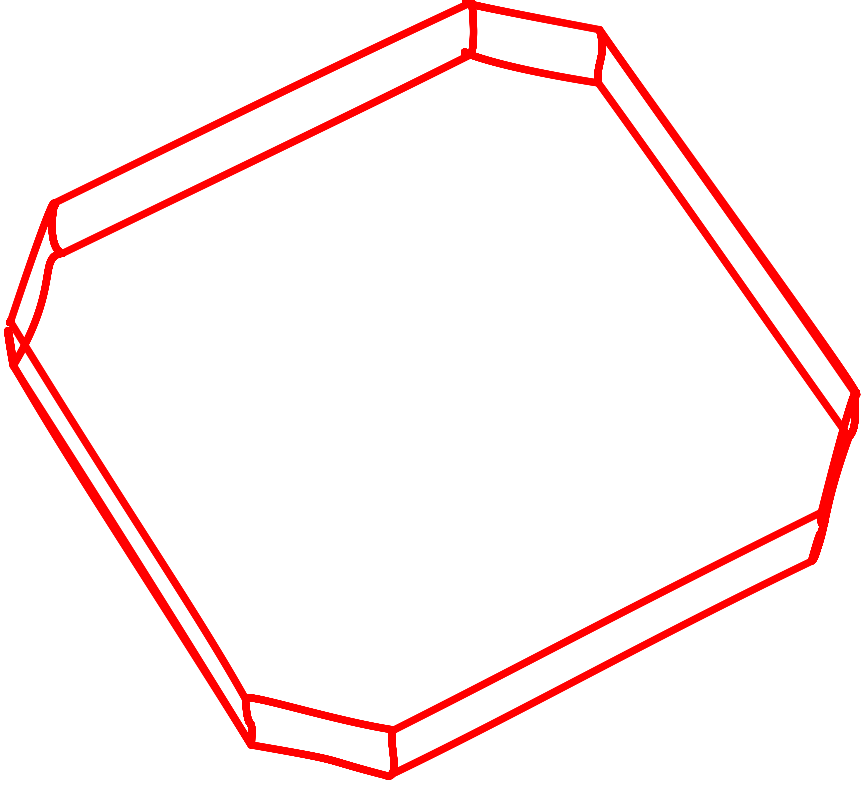}}
			
		\end{minipage}
		\begin{minipage}{0.137\linewidth}
			\centerline{Ours (Edge)}
			\vspace{2pt}
			\centerline{\includegraphics[width=\textwidth]{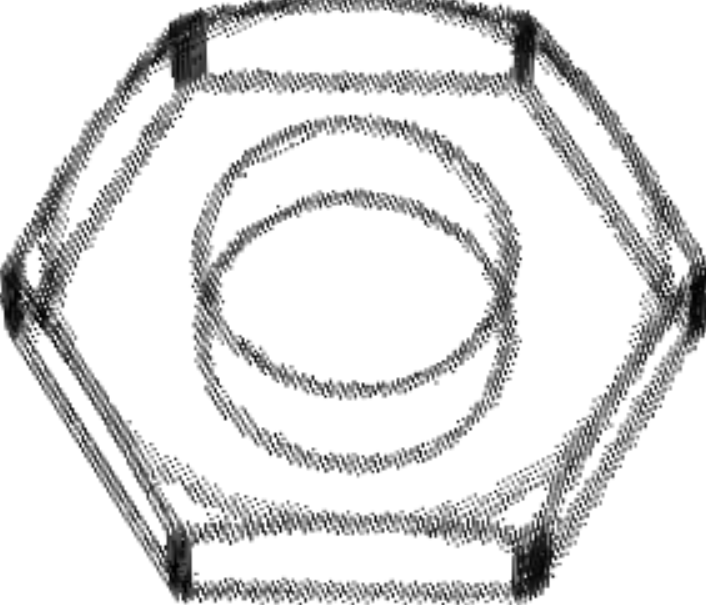}}
			\vspace{1.5pt}
			\centerline{\includegraphics[width=\textwidth]{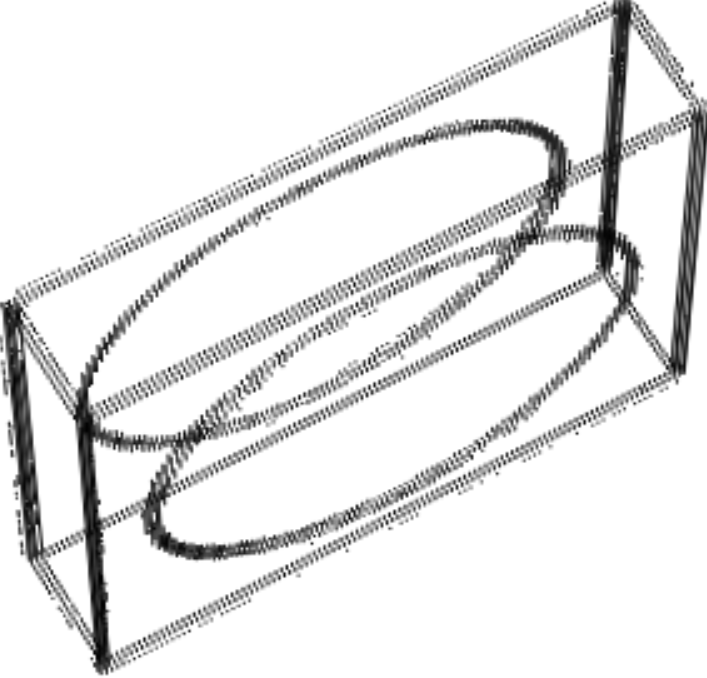}}
			\vspace{1.5pt}
			\centerline{\includegraphics[width=\textwidth]{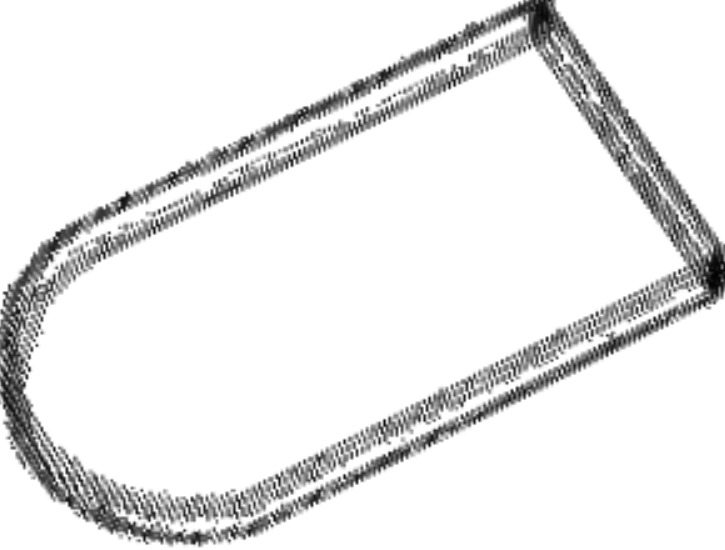}}
			\vspace{1.5pt}
			\centerline{\includegraphics[width=\textwidth]{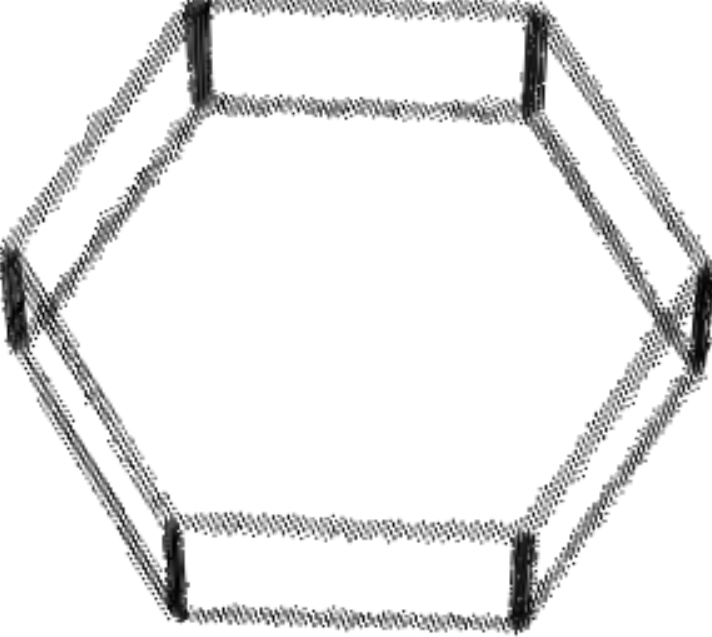}}
			\vspace{1.5pt}
			\centerline{\includegraphics[width=\textwidth]{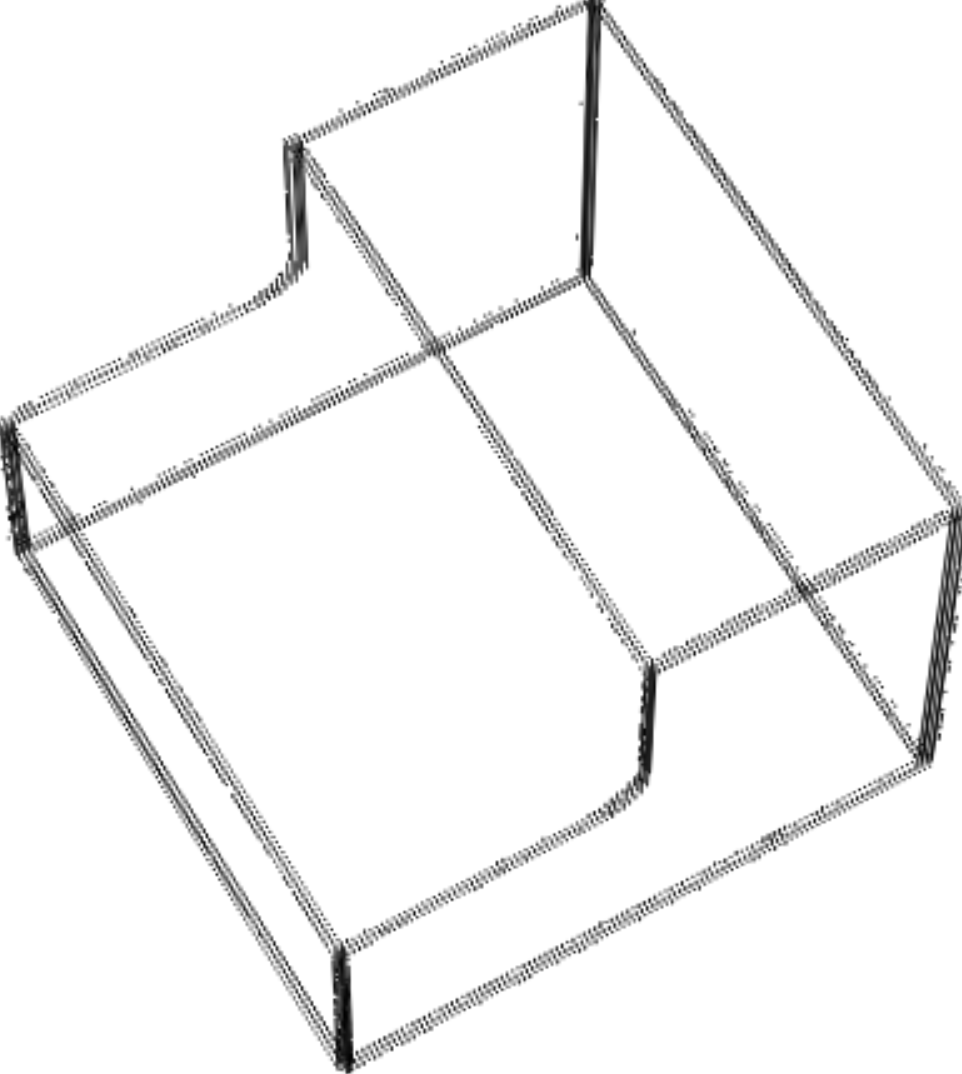}}
			\vspace{1.5pt}
			\centerline{\includegraphics[width=\textwidth]{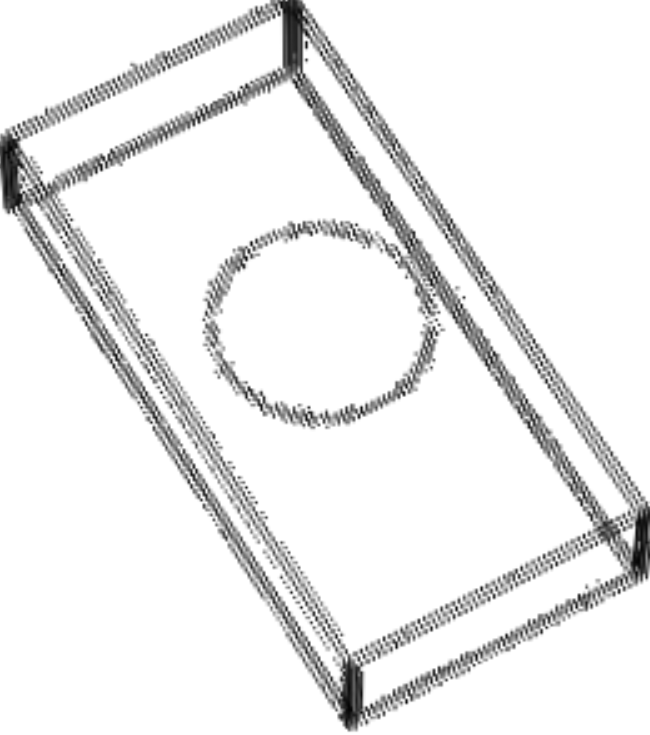}}
			\vspace{1.5pt}
			\centerline{\includegraphics[width=\textwidth]{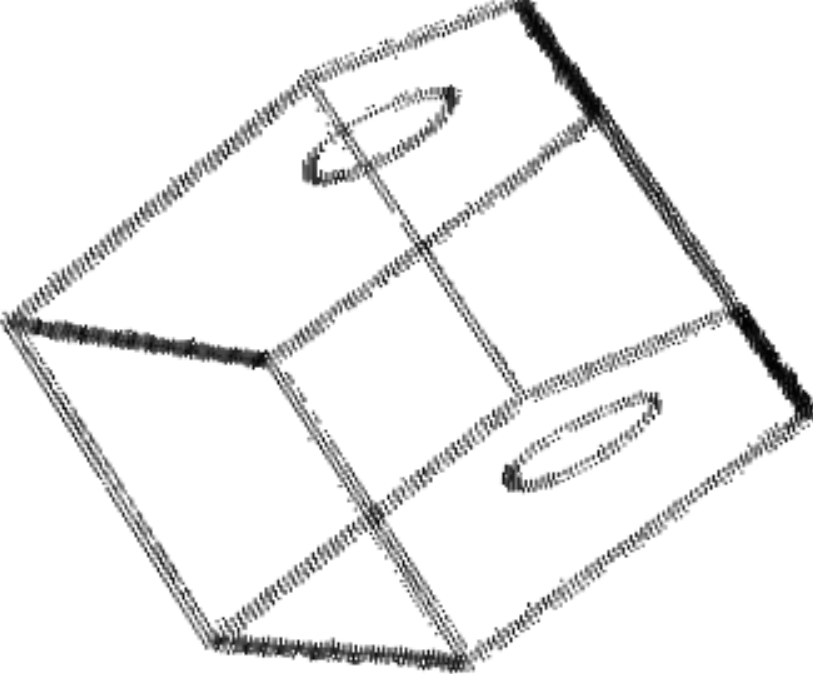}}
			\vspace{1.5pt}
			\centerline{\includegraphics[width=\textwidth]{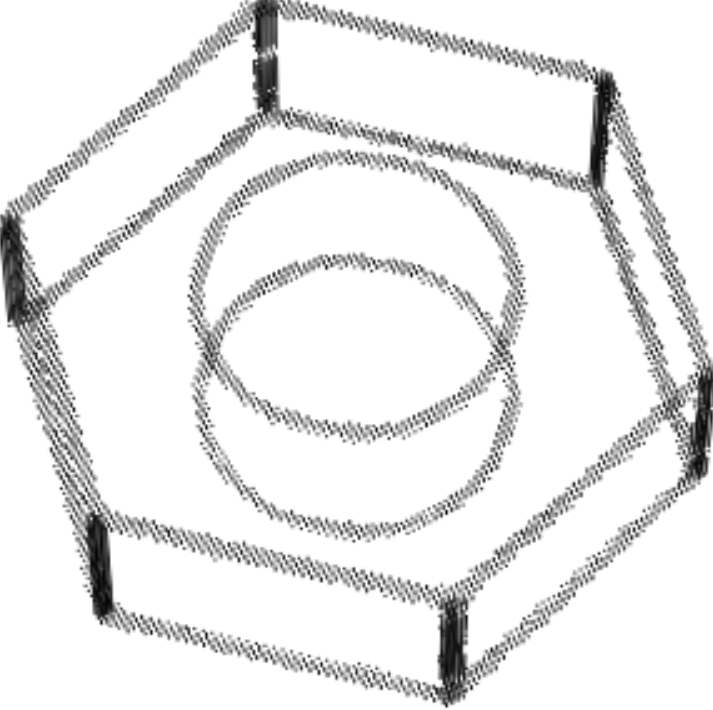}}
			\vspace{1.5pt}
			\centerline{\includegraphics[width=\textwidth]{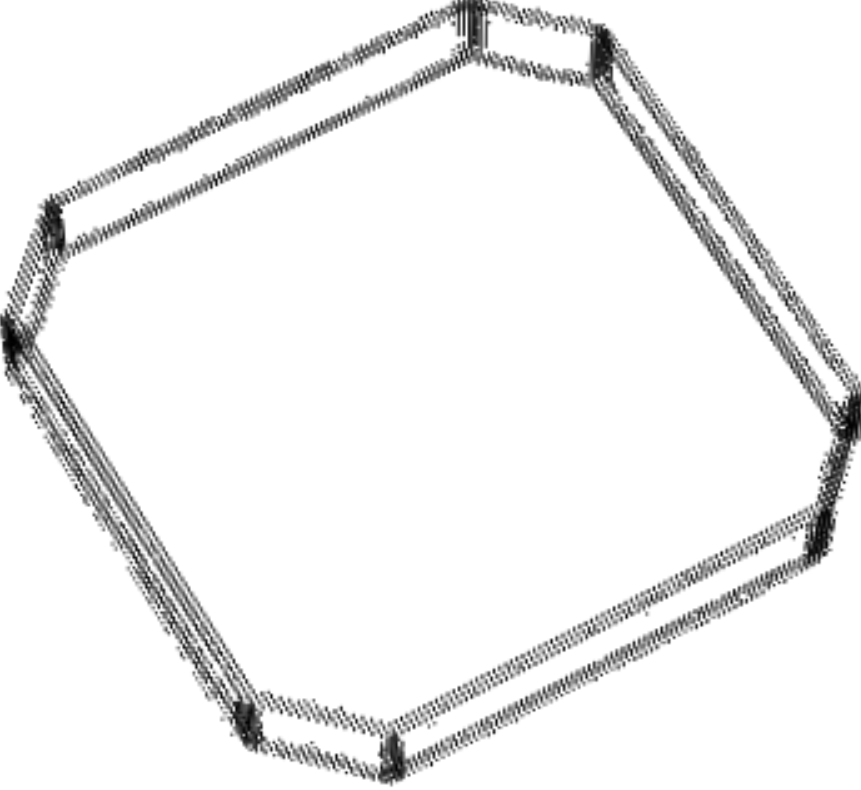}}
			
		\end{minipage}
		\begin{minipage}{0.137\linewidth}
			\centerline{GT}
			\vspace{2pt}
			\centerline{\includegraphics[width=\textwidth]{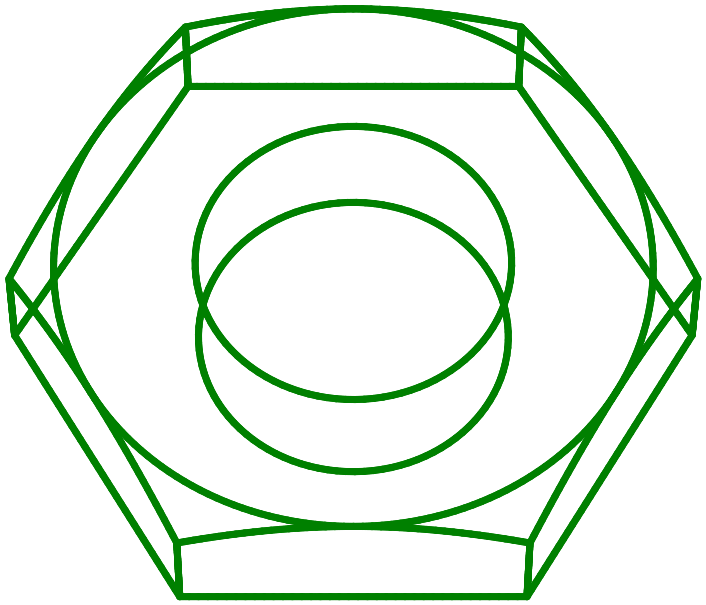}}
			\vspace{1.5pt}
			\centerline{\includegraphics[width=\textwidth]{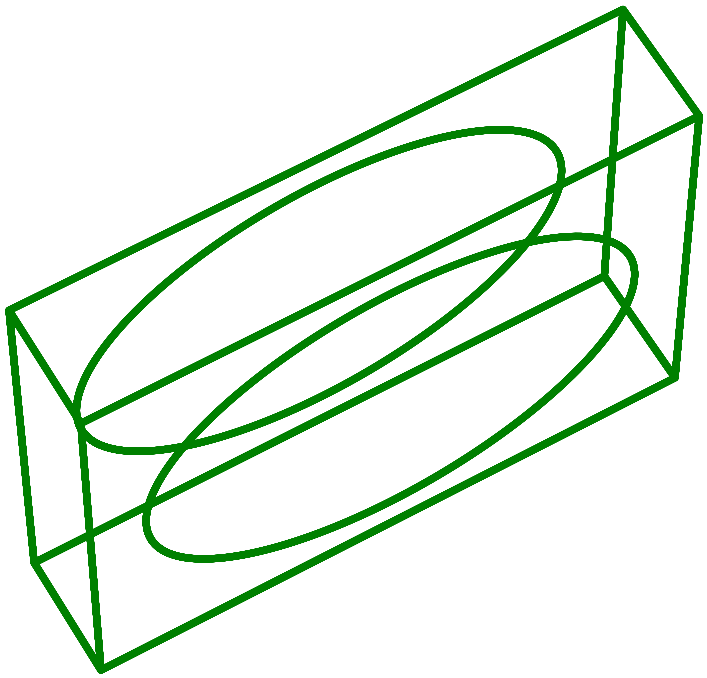}}
			\vspace{1.5pt}
			\centerline{\includegraphics[width=\textwidth]{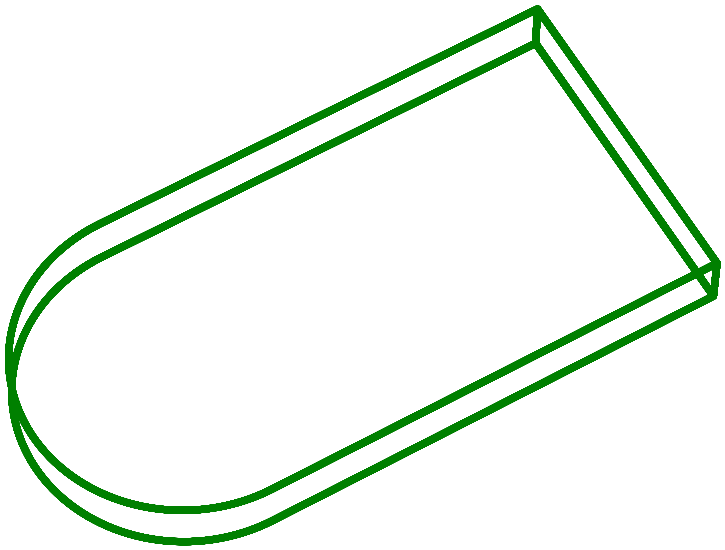}}
			\vspace{1.5pt}
			\centerline{\includegraphics[width=\textwidth]{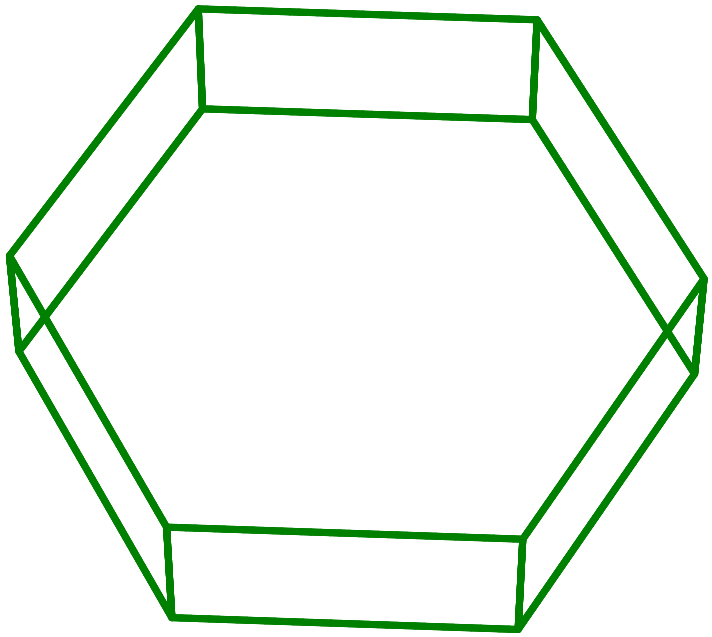}}
			\vspace{1.5pt}
			\centerline{\includegraphics[width=\textwidth]{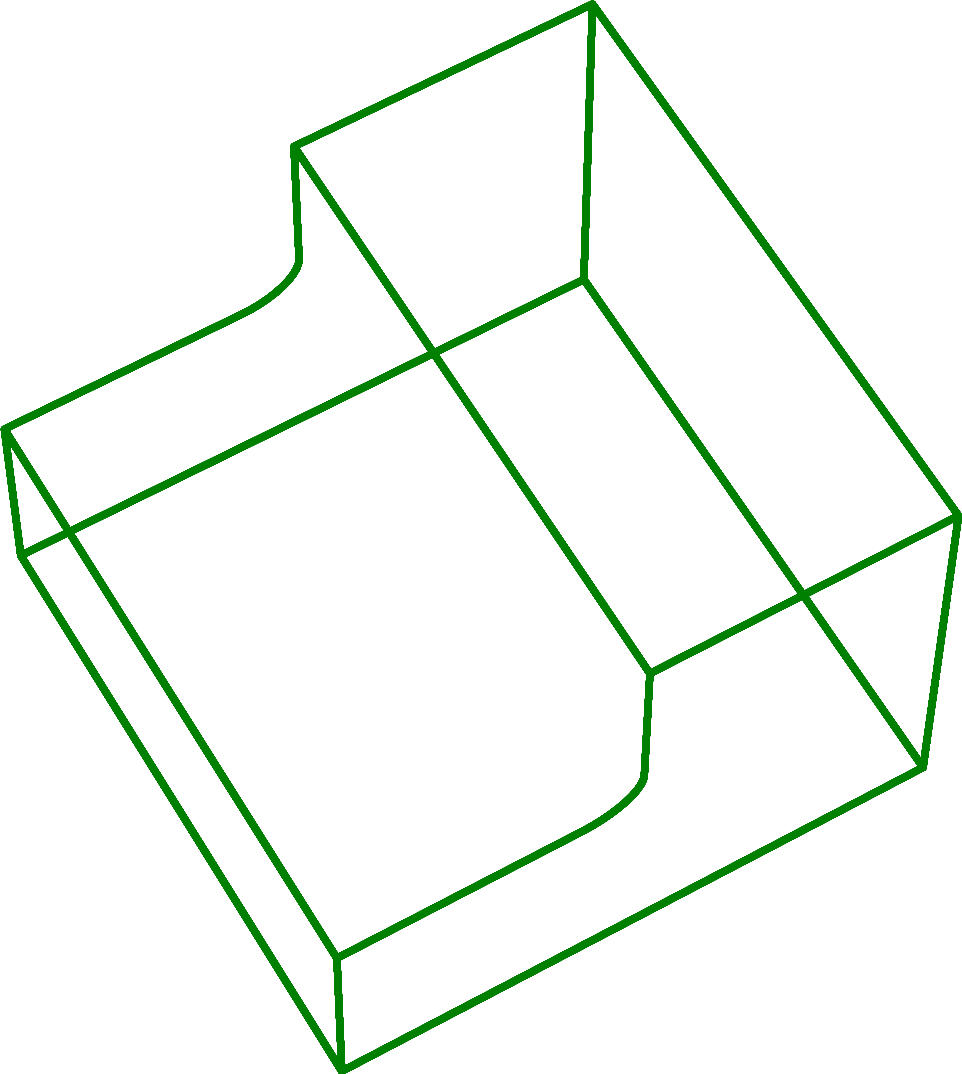}}
			\vspace{1.5pt}
			\centerline{\includegraphics[width=\textwidth]{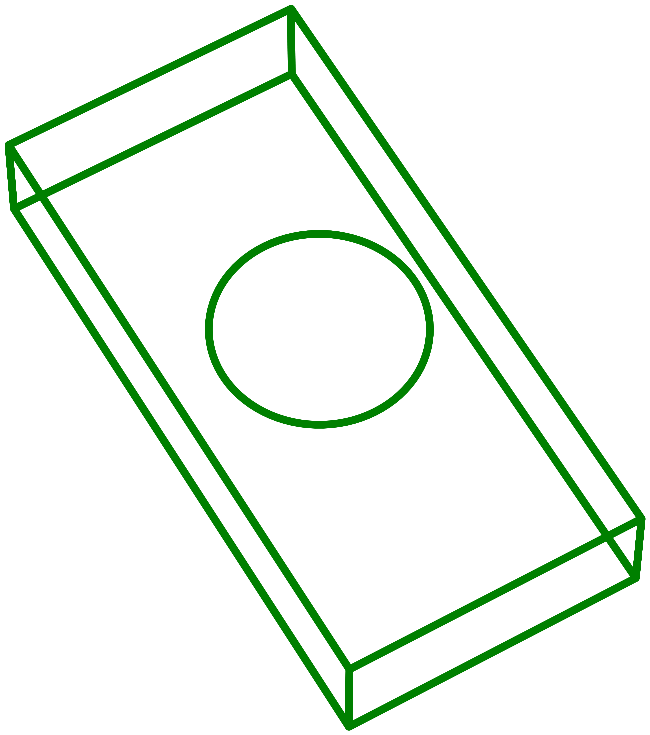}}
			\vspace{1.5pt}
			\centerline{\includegraphics[width=\textwidth]{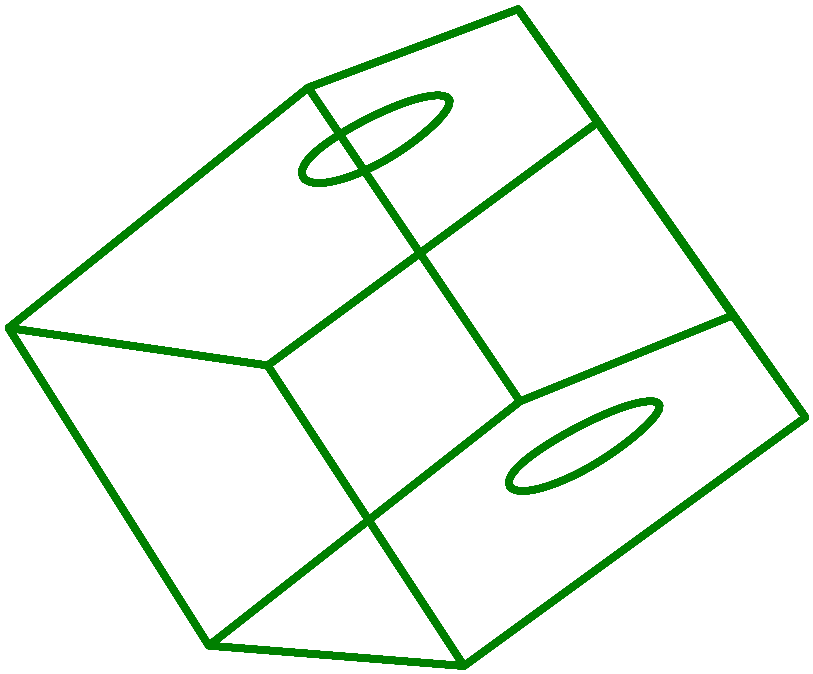}}
			\vspace{1.5pt}
			\centerline{\includegraphics[width=\textwidth]{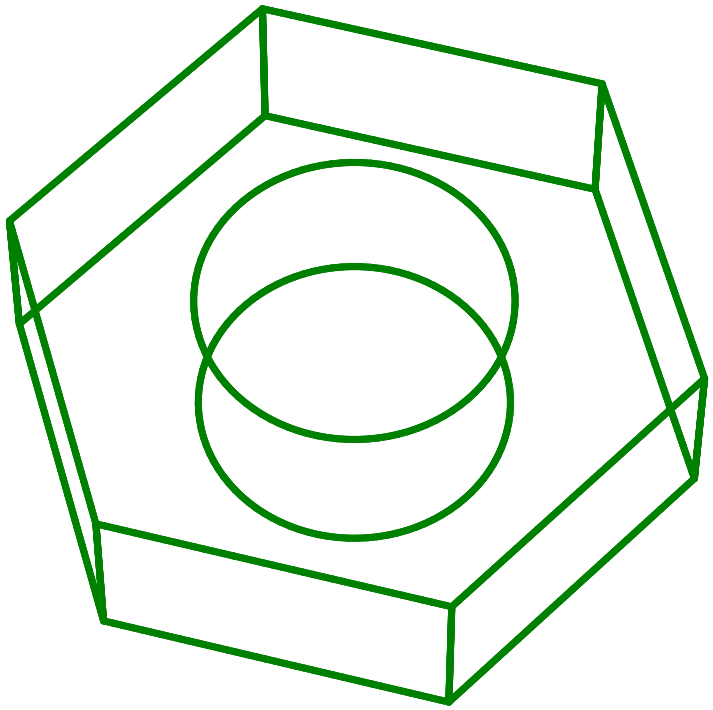}}
			\vspace{1.5pt}
			\centerline{\includegraphics[width=\textwidth]{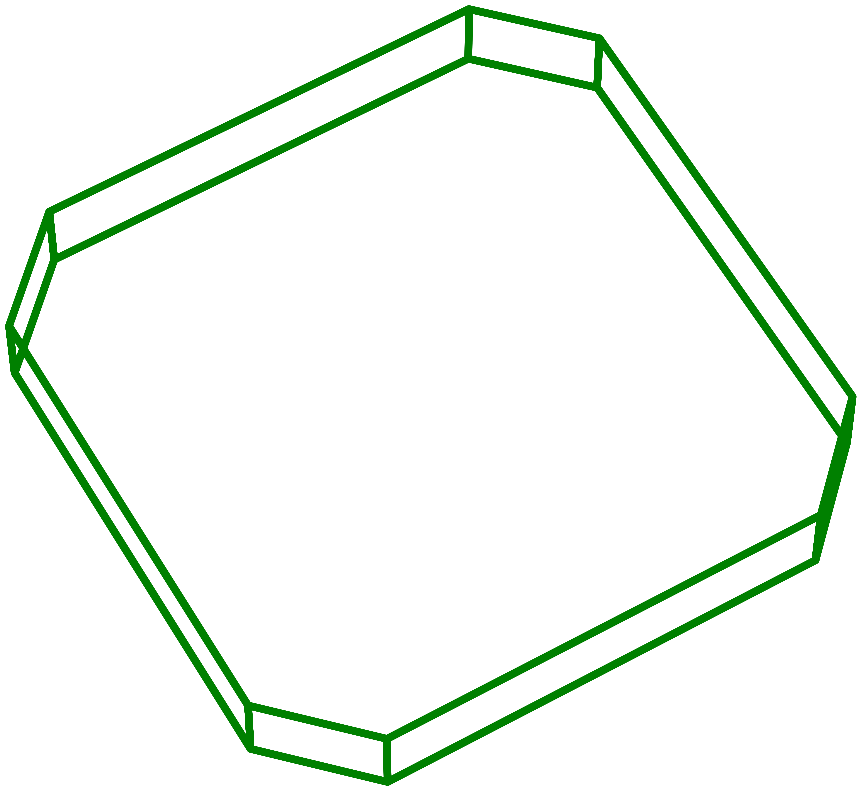}}

		\end{minipage}
	\end{center}
	\caption{More qualitative comparisons against other state-of-the-arts. From left to right, we present the rendered image, the result curves of PIE-NET, PC2WF, DEF, our reconstructed curves, our 3D edge points obtained from edge densities, and the ground truth edges.}
	
	\label{fig:qualitative_results}
\end{figure*}

We provide more qualitative comparisons with state-of-the-art methods of parametric curve reconstruction, including PIE-NET~\cite{wang2020pie}, PC2WF~\cite{liu2021pc2wf} and DEF~\cite{matveev2022def}. The results are illustrated in Fig.~\ref{fig:qualitative_results}.

\section{Limitations}
To foster additional works in this field, we briefly demonstrate several limitations of NEF, which are also potential directions for future work. 

\textit{Training speed.} 
Currently, it takes about one hour for NEF to train each model with 50 views, one can reduce the number of views to speed up with minor performance drops in most cases, as shown in~\ref{sec:ablation}. Also, the edge densities in spatial positions are highly sparse and could be accelerated by decreasing the samples along rays or integrating other voxel-based NeRF works for speedup. The coarse and fine optimization stages cost about 30 and 4 seconds on average, respectively.

\textit{Textured objects.} 
3D edges exactly lie in areas where normal changes sharply, while 2D edges also contain other edge types (e.g. shadow, surface texture). Objects with rich textures could bring much noise on 2D edge maps and consequently influence extracting 3D edge points and reconstructing curves. Those noisy edges could be suppressed from both the image level (classify which edge pixel is caused by texture discontinuity) and NEF level (recognize texture edge densities by locating object surfaces).

\textit{Edges inside the object.} 
We cannot detect unseen edges hidden inside the object from only 2D images, and thus cannot reconstruct the corresponding curves. This is a natural drawback of our method, and could be tackled by integrating extra 3D cues (e.g. point cloud, mesh, shape prior).

\end{appendix}

\end{document}